\pdfoutput=1
\documentclass[11pt]{article}
\usepackage{acl}

\usepackage{times}
\usepackage{latexsym}
\usepackage{amsmath,amssymb,amsthm,mathtools}
\usepackage{booktabs,multirow,makecell}
\usepackage{graphicx}
\usepackage{microtype}
\usepackage{inconsolata}
\usepackage{enumitem}
\usepackage{url}
\usepackage{algorithm}
\usepackage{algpseudocode}
\usepackage{xcolor}
\usepackage{tikz}
\usepackage{hyperref}
\usepackage{caption}
\usepackage{subcaption}
\usepackage{float}
\usepackage{xspace}
\usepackage[T1]{fontenc}
\usepackage[utf8]{inputenc}
\usepackage{listings}
\newif\ifannotations
\annotationsfalse
\ifannotations
 \newcommand{\mg}[1]{\textcolor{red}{MG:#1}}
 \newcommand{\torun}[1]{\textcolor{red}{\textbf{[TODO: #1]}}}
\else
 \newcommand{\mg}[1]{}
 \newcommand{\torun}[1]{}
\fi

\newcommand{\SCULPT}{{\small SCULPT}\xspace}
\newcommand{\protegi}{{\small ProTeGi}\xspace}
\newcommand{\OPRO}{{\small OPRO}\xspace}
\newcommand{\Distill}{{\small DISTILL}\xspace}
\newcommand{\CRAFT}{{\small CRAFT}\xspace}
\newcommand{\WPRO}{{\small WPRO}\xspace}
\newcommand{\Evo}{{\small EvoPrompt}\xspace}
\newcommand{\LLMLingua}{{\small LLMLingua}\xspace}
\DeclareMathOperator*{\argmax}{arg\,max}

\newcommand{\crafticon}{\tikz[baseline=2pt, scale=1.2, line cap=round, line join=round]{%
 \fill[orange!75!yellow!95, draw=black, line width=0.3pt]
 (0.030,0.090) -- (0.090,0.030) -- (0.240,0.180) -- (0.180,0.240) -- cycle;
 \fill[orange!35!yellow!55, draw=black, line width=0.3pt]
 (0.180,0.240) -- (0.240,0.180) -- (0.310,0.310) -- cycle;
 \fill[gray!25!black, draw=black, line width=0.3pt]
 (0.252,0.288) -- (0.288,0.252) -- (0.310,0.310) -- cycle;
 \draw[line width=1.3pt, black] (0.36,0.30) -- (0.554,0.088);
 \draw[line width=1.3pt, black] (0.58,0.30) -- (0.386,0.088);
 \fill[black] (0.47,0.18) circle (0.024);
 \draw[line width=1.0pt, black, fill=white] (0.36,0.06) circle (0.050);
 \draw[line width=1.0pt, black, fill=white] (0.58,0.06) circle (0.050);
}}
\title{\crafticon\,\CRAFT: Cost-aware Refinement And Front-aware Tuning of Prompts}

\author{\mdseries%
  Shanu Kumar\textsuperscript{1}, Shubhanshu Khandelwal\textsuperscript{2}, Akhila Yesantarao Venkata\textsuperscript{2}, \\
  Parag Agrawal\textsuperscript{2}, Yova Kementchedjhieva\textsuperscript{1}, Manish Gupta\textsuperscript{2} \\
  \textsuperscript{1}MBZUAI \quad \textsuperscript{2}Microsoft \\
  {\tt\small \{shanu.kumar,yova.kementchedjhieva\}@mbzuai.ac.ae} \\
  {\tt\small \{shukhand,akyesant,paragag,gmanish\}@microsoft.com}
}

\begin{document}
\maketitle

\begin{abstract}
Prompts tuned for accuracy often grow long, raising inference cost on every model call. The best accuracy-cost trade-off depends on the task and the budget, so prompt optimization is a search over the \emph{Pareto front} of accuracy and prompt-token cost rather than for one prompt. The usual shortcut, collapsing the objectives into a weighted sum, fixes the trade-off weight before search and often recovers only a narrow region of the front, a failure we call \emph{scalarization collapse}. We present \textbf{\CRAFT} (Cost-aware Refinement And Front-aware Tuning), a Pareto-front prompt optimizer that treats target-LLM validation calls as the scarce resource and allocates them to candidates near the optimistic candidate front. Each round, complementary accuracy-oriented and cost-oriented generators propose edits, Pareto-gap acquisition spends the per-round validation budget, and NSGA-II retention keeps a spread-out population. Across six classification and reasoning benchmarks, \CRAFT's retained fronts reach both high-accuracy and low-cost regions, while accuracy-only, cost-only, and weighted-sum baselines each concentrate in narrower regions. The accuracy-cost trade-off becomes a post-search choice, not a pre-search weight.
\end{abstract}

\section{Introduction}
\label{sec:intro}
Large language models (LLMs) are adapted to downstream tasks by tuning the natural-language prompt fed to the model \citep{ramnath2025systematic}. Longer, more structured prompts cost more per call, and small edits change predictions disproportionately \citep{10.1162/tacl_a_00324,zhao2021calibrate}, motivating automatic prompt optimization.

Prior work splits along two paths. Accuracy-oriented optimizers, e.g., \OPRO \citep{yang2023large}, \protegi \citep{pryzant-etal-2023-automatic}, \SCULPT \citep{kumar-etal-2025-sculpt}, and \Evo \citep{guo2024connecting}, maximize task accuracy; prompt-compression methods, e.g., Gist tokens \citep{mu2023learning}, \LLMLingua and \LLMLingua-2 \citep{jiang-etal-2023-llmlingua,pan-etal-2024-llmlingua}, reduce token count to lower per-call inference cost \citep{chen2024frugalgpt}. Both lines treat a prompt as good or bad along one axis, yet two prompts of similar accuracy but different lengths are not equivalent: each extra token adds to inference cost on every model call, so accuracy and cost are coupled.


\begin{figure}[!t]
\centering
\includegraphics[width=0.95\linewidth]{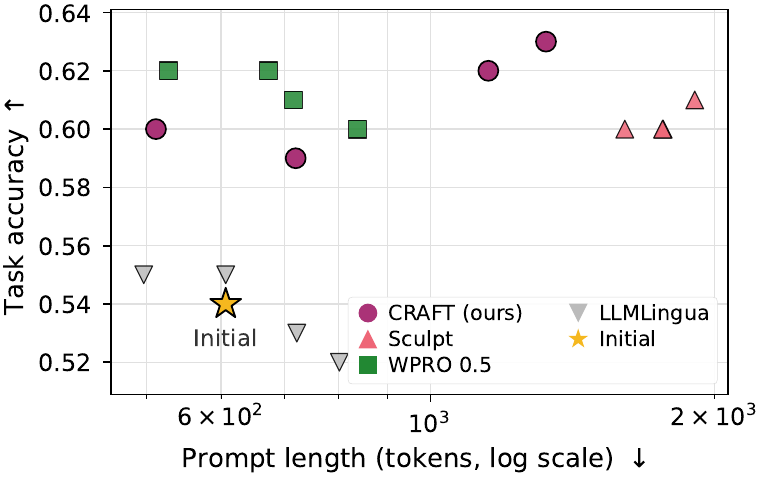}
\caption{Accuracy-cost trade-off on BeaverTails. \SCULPT (accuracy-only), \LLMLingua (cost-only), and {\WPRO}$_{0.5}$ (weighted-sum) each collapse to a narrow region of the front. \CRAFT (ours) covers both high-accuracy and low-token regions of the front.}
\label{fig:motivation}
\end{figure}

This creates a multi-objective prompt-optimization problem. The objectives often conflict: higher accuracy tends to require more detailed prompts, while lower cost favors concise ones. The relationship between prompt text and both accuracy and cost is non-convex, and the prompt space is discrete \citep{wen2024hard,10.1162/tacl_a_00324,zhao2021calibrate}, so the trade-off surface contains regions that a fixed scalar reward misses \citep{das1997closer}. We instead optimize the accuracy-cost Pareto front directly, defined as the set of prompts for which no other prompt is both more accurate and shorter.

The algorithmic challenge is not merely choosing among evaluated trade-offs, but deciding what to evaluate: each candidate's accuracy requires target-LLM validation calls, so the optimizer must spend a per-round validation budget before the candidate's true front position is known. Optimizing a fixed scalar reward before search often recovers only a narrow region of the front, the \emph{scalarization collapse} above. We propose \CRAFT (Cost-aware Refinement And Front-aware Tuning), a front-aware optimization loop: the refiner and condenser generate edits with complementary accuracy and cost biases; Pareto-gap acquisition \citep{srinivas2009gaussian,NEURIPS2021_11704817} allocates the per-round validation budget to candidates near the optimistic candidate front; and Non-dominated Sorting Genetic Algorithm II (NSGA-II) retention \citep{deb2002fast} keeps a diverse validated population for the next round. Figure~\ref{fig:motivation} illustrates the contrast: accuracy-only, cost-only, and weighted-sum baselines each concentrate on a narrow region of the front, while \CRAFT covers both ends.

Our contributions are: (i) we formalize cost-aware prompt optimization as budgeted Pareto-front search over accuracy and prompt-token cost, where target-LLM validation calls govern which candidate trade-offs can be observed; (ii) we propose \textbf{\CRAFT},\footnote{Code: \url{https://github.com/Sshanu/CRAFT}} a front-aware framework that separates candidate generation from validation allocation, using complementary accuracy-oriented and cost-oriented generators, Pareto-gap acquisition, and NSGA-II retention as one budgeted loop; (iii) we show that accuracy-only, cost-only, and weighted-sum baselines concentrate in narrow regions of the front, while \CRAFT retains feasible prompts across high-accuracy and low-cost parts of the front on six benchmarks and transfers across optimizer LLMs.

\begin{figure*}[!tb]
\centering
\includegraphics[width=0.9\linewidth]{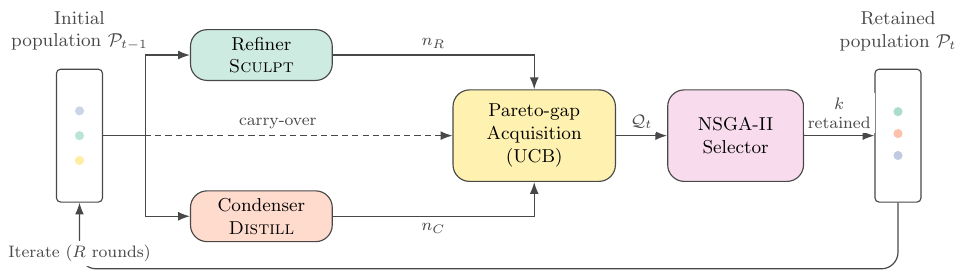}
\caption{\CRAFT framework. Each round, the refiner and condenser generate new candidates from the previous population $\mathcal{P}_{t-1}$; Pareto-gap acquisition validates them under the per-round validation budget; the NSGA-II selector retains a non-dominated, spread-out population $\mathcal{P}_t$ for the next round. The loop repeats for $R$ rounds.}
\label{fig:framework}
\end{figure*}

\section{Related Work}
\label{sec:related}

\paragraph{Prompt Refinement.}
Prompt optimizers fall into four families, based on the optimization signal they use: (1) LLM-edit methods rewrite prompts via another LLM \citep{yang2023large,pryzant-etal-2023-automatic,kumar-etal-2025-sculpt}, (2) gradient or program-based methods that propagate signal through differentiable or symbolic graphs \citep{yuksekgonul2024textgrad,khattab2024dspy}, (3) RL-based methods \citep{deng-etal-2022-rlprompt}, and (4) evolutionary mutation-and-selection methods \citep{guo2024connecting,pmlr-v235-fernando24a}. Most operate on raw text tokens, which makes large structural edits hard to express. A small subset of refiners edit at the compile-time or tree-level abstraction \citep{schnabel-neville-2024-symbolic,kumar-etal-2025-sculpt}. \CRAFT builds on this structure-aware line, adopting \SCULPT \citep{kumar-etal-2025-sculpt} as its default refiner (\S\ref{sec:refiner}) for its accuracy-oriented strength and treating it as swappable, with drop-in replacements tested in Table~\ref{tab:ablation}.

\paragraph{Prompt Compression.}
Token-level compressors drop low-information surface tokens to reduce length \citep{mu2023learning,jiang-etal-2023-llmlingua,pan-etal-2024-llmlingua}. Because they operate below the level of instructions or constraints, aggressive compression can remove task-critical content; this limits their value as a cost-oriented module inside a Pareto-front loop. A structure-aware compressor that edits the parsed prompt tree could avoid this failure mode, but we are not aware of prior work in this direction; we introduce one (\Distill, \S\ref{sec:condenser}) as \CRAFT's cost-oriented generator.

\paragraph{Multi-Objective Prompt Optimization.}
Existing multi-objective prompt optimizers \citep{jafari-etal-2024-morl,resendiz2024mopo,zhao2025pareto,yang-li-2023-instoptima,agrawal2026gepa} optimize multi-objective rewards, but they do not study front construction when each candidate's accuracy must be estimated with target-LLM validation calls and a fixed budget can cover only part of the discrete prompt set. \CRAFT targets that budgeted front-construction setting for accuracy versus prompt-token cost. The closest cost-aware optimizer is CAPO (Cost-Aware Prompt Optimization) \citep{zehle2025capo}; we adapt its fixed-weight scalarization as our \WPRO baseline; the resulting scalarization collapse is shown in Figure~\ref{fig:motivation}.

\paragraph{Multi-Objective Bayesian Optimization and Cost-Aware Inference.}
Multi-objective Bayesian optimization \citep{knowles2006parego,srinivas2009gaussian,NEURIPS2021_11704817} frames our problem but assumes numerical design spaces and Gaussian-process surrogates, neither available for discrete text; an acquisition rule on partial-validation scores is needed instead. Cost-aware LLM inference \citep{chen2024frugalgpt,yue2024large} routes inputs across models at inference time and is orthogonal to prompt optimization.

\section{The \CRAFT Framework}
\label{sec:framework}

\CRAFT maintains a small population of prompts and, round by round, reshapes it into an approximation of the accuracy-cost Pareto front. Each round has three steps: a structure-aware refiner and condenser propose new candidates (\S\ref{sec:refiner}, \S\ref{sec:condenser}); a Pareto-gap acquisition function spends the per-round validation budget on the candidates near the optimistic candidate front (\S\ref{sec:approximator}); and a population selector keeps $k$ prompts spread across the front for the next round (\S\ref{sec:selector}). Figure~\ref{fig:framework} shows the loop.

\subsection{Problem Formulation}
\label{sec:problem}

We cast prompt optimization as a bi-objective search over a task-accuracy score and a prompt-token cost. Let $\mathcal{D}_{\mathrm{val}}$ and $\mathcal{D}_{\mathrm{test}}$ be the validation and held-out test sets, of sizes $N_{\mathrm{val}}$ and $N_{\mathrm{test}}$. For a prompt $P$ and a subset $\mathcal{V}\subseteq\mathcal{D}_{\mathrm{val}}$, $p(P;\mathcal{V})$ is the task score on $\mathcal{V}$. For a candidate $P_i$ we write $p_i=p(P_i;\mathcal{D}_{\mathrm{val}})$ for its full-validation score (the objective that search ultimately optimizes), and $c_i=c(P_i)$ for its cost, the number of prompt tokens, excluding example inputs and model outputs. During search $p_i$ is estimated from partial validation (\S\ref{sec:approximator}), while held-out test scores are reserved for final reporting (Appendix~\ref{app:datasets}). The two objectives are in tension, so no single prompt is uniformly best. A prompt $P_i$ \emph{dominates} $P_j$, written $P_i \succ P_j$, if it is no worse on both objectives and strictly better on one: $p_i \ge p_j$ and $c_i \le c_j$, with at least one strict inequality. The prompts of a set $\mathcal{C}$ that no other prompt dominates form its \emph{Pareto-optimal set} $\mathcal{C}^{*}$, and \CRAFT's goal is to approximate $\mathcal{C}^{*}$ over the discrete space of LLM prompts under a fixed budget of LLM calls per round.

\CRAFT builds this approximation iteratively, starting from a single hand-written prompt. Round $t$ opens with a retained population $\mathcal{P}_{t-1}$ of $k$ prompts. The refiner and condenser expand it into a candidate set $\mathcal{Q}_t$; the acquisition function validates a budgeted subset of $\mathcal{Q}_t$; and the selector keeps $k$ validated prompts as the next population $\mathcal{P}_t$. After $R$ rounds, $\mathcal{P}_R$ is the returned front approximation.

\subsection{Refiner}
\label{sec:refiner}
\CRAFT draws candidates from two modules with opposite biases. Optimizing accuracy alone inflates prompts; optimizing cost alone strips away task signal. An accuracy-oriented \emph{refiner} (this subsection) targets higher task accuracy, and a cost-oriented \emph{condenser} (\S\ref{sec:condenser}) targets lower prompt-token cost; keeping both in the population lets the selector trade the two off.
The refiner is any accuracy-oriented prompt optimizer compatible with the per-round budget interface; \SCULPT, \OPRO, and \Evo are all interchangeable in this role. We use \SCULPT as the default for two reasons: prior work shows it generalizes across a broader set of benchmarks than \OPRO \citep{kumar-etal-2025-sculpt}, and its tree-level edits are precise and stable. \SCULPT's critic-actor scaffold (Appendix~\ref{app:modules}, Figure~\ref{fig:actor_critic}) parses the prompt into a hierarchical tree of sections, rules, and examples; a critic-actor pair edits that tree by reordering, elaborating, or clarifying nodes, emitting up to $n_R$ variants per prompt each round. Prompt templates and hyperparameters for \SCULPT, \OPRO, and \Evo are listed in Appendix~\ref{app:modules}.

\subsection{Condenser}
\label{sec:condenser}
The condenser lowers prompt-token cost while preserving task-relevant instructions. Our default \textbf{\Distill} reuses \SCULPT's critic-actor scaffold for its precision and stability, making it structure-aware by design. Token-level compressors such as \LLMLingua-2 drop low-information surface tokens and can discard task-critical instructions; \Distill instead edits the parsed prompt tree: the critic flags redundant or verbose substructures, and the actor applies the tree-level edits listed in Table~\ref{tab:distill_actions}, which update, delete, or merge nodes. Each round, every prompt is condensed at $n_C$ compression ratios sampled from a fixed interval (Appendix~\ref{app:modules}).

\begin{figure}[!t]
\centering
\includegraphics[width=\linewidth]{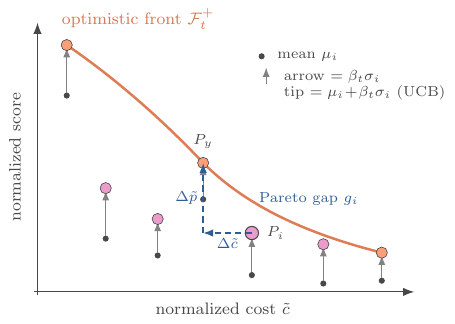}
\caption{Pareto-gap acquisition geometry. Dots mark candidate means at normalized cost $\tilde{c}_i$; grey arrows extend means by $\beta_t\sigma_{i,t}$ to UCB tips at optimistic score $\tilde{p}_{i,t}$. Orange tips form the optimistic candidate front $\mathcal{F}^{+}_t$; pink tips lie below it. For candidate $P_i$, the dashed L-shape to nearest front tip $P_y$ has length $g_i=\Delta\tilde{c}+\Delta\tilde{p}$.}
\label{fig:performance_approx}
\end{figure}

\subsection{Pareto-Gap Acquisition Function}
\label{sec:approximator}

Candidate generation by the refiner and condenser uses optimizer-LLM calls, but it is cheaper than scoring many candidates with target-LLM calls over validation examples. Scoring every prompt in $\mathcal{Q}_t$ on all of $\mathcal{D}_{\mathrm{val}}$ would consume the per-round validation budget many times over. \CRAFT therefore validates a budgeted subset each round and scores each selected candidate on the current validation subset. We partition $\mathcal{D}_{\mathrm{val}}$ once into $T$ subsets $\{\mathcal{V}_j\}_{j=1}^{T}$ by $k$-means clustering; round $t$ uses subset $\mathcal{V}_{\pi(t)}$, cycling through them with $\pi(t)=((t-1)\bmod T)+1$. This shrinks per-validation cost by a factor of $T$; subset sizes and the rotation schedule are detailed in Appendix~\ref{app:val_subsets}.

Because the true Pareto front is unknown during search, \CRAFT spends the per-round validation budget \emph{optimistically}. For each candidate $P_i$ it keeps a running mean $\mu_{i,t}$ and standard deviation $\sigma_{i,t}$ of the per-example scores observed so far (formulas in Appendix~\ref{app:primer_ucb}), and combines them into an Upper Confidence Bound (UCB) score $p^+_{i,t}=\mu_{i,t}+\beta_t\sigma_{i,t}$. A candidate with few observations has a large $\sigma_{i,t}$, so it is judged as if its true score sat at the optimistic end of its plausible range. The weight $\beta_t$ falls linearly from near $1$ in the first round to $0.5$ in the last, tilting the search from exploration toward exploitation. A brand-new candidate has no observations at all (we initialize $\mu_{i,t}{=}0$, $\sigma_{i,t}{=}1$), so it receives the default exploratory score $\beta_t$. We denote by $\mathcal{S}_{i,t}$ the observation set of $P_i$, i.e., the validation examples on which it has been scored through round $t$.

\begin{algorithm}[!t]
\footnotesize
\caption{Pareto-Gap UCB Acquisition}
\label{alg:pgucb}
\begin{algorithmic}[1]
\Require Current candidates $\mathcal{Q}_t$, means $\{\mu_{i,t}\}$, empirical stds $\{\sigma_{i,t}\}$, costs $\{c_i\}$, observation sets $\{\mathcal{S}_{i,t}\}$, validation batch size $B$, round $t$, subset count $T$
\Ensure Evaluation batch $\mathcal{E}_t$ with $|\mathcal{E}_t| \le B$
\State Compute $\beta_t$ and $p^+_{i,t}=\mu_{i,t}+\beta_t\sigma_{i,t}$ for all $P_i \in \mathcal{Q}_t$
\State Min-max normalize $\{p^+_{i,t}\}$ and $\{c_i\}$ over $\mathcal{Q}_t$ to obtain $(\tilde{p}_{i,t}, \tilde{c}_i)$
\State $\mathcal{F}^{+}_t \gets \textsc{ParetoFront}\big(\{(\tilde{p}_{i,t}, \tilde{c}_i) : P_i \in \mathcal{Q}_t\}\big)$
\For{each candidate $P_i \in \mathcal{Q}_t$}
 \State $g_i \gets \min\limits_{P_y \in \mathcal{F}^{+}_t}\big[\max(0,\tilde{p}_{y,t}-\tilde{p}_{i,t})+\max(0,\tilde{c}_i-\tilde{c}_y)\big]$
\EndFor
\State Sort $\mathcal{Q}_t$ by ascending $g_i$, then descending $p^+_{i,t}$, then ascending $c_i$, then ascending $|\mathcal{S}_{i,t}|$, then stable prompt ID
\State \Return the first $B$ candidates as $\mathcal{E}_t$
\end{algorithmic}
\end{algorithm}

The UCB score ranks candidates by their optimistic estimate alone; \CRAFT also favors the ones that would extend the \emph{front}. It pairs each UCB score with the candidate's cost and min-max normalizes both over the current candidate set, placing $P_i$ at $(\tilde{p}_{i,t}, \tilde{c}_i)$. In Figure~\ref{fig:performance_approx}, the non-dominated points, drawn as orange UCB tips, form a temporary, optimistic candidate front $\mathcal{F}^+_t$. The \emph{Pareto gap} $g_i$ then measures how far an off-front candidate sits from $\mathcal{F}^+_t$, as the Manhattan distance to the nearest front point within the dominated quadrant, shown as a dashed L-shape from $P_i$ to its nearest front tip $P_y$:
\[
g_i = \min_{y\in\mathcal{F}^+_t}\!\big[\,\max(0,\tilde{p}_{y,t}\!-\!\tilde{p}_{i,t}) + \max(0,\tilde{c}_i\!-\!\tilde{c}_y)\,\big].
\]
A candidate already on the front has $g_i = 0$; the smaller its gap, the more likely validating it is to improve the retained front. Brand-new candidates all share the UCB score $\beta_t$, so among them the gap reduces to a preference for cheaper prompts until observations accrue. In Algorithm~\ref{alg:pgucb}, we rank $\mathcal{Q}_t$ by ascending gap and take the $B$ smallest as the round's validation batch $\mathcal{E}_t$. We set $B = 2k$, twice the retained population, letting the selector pick a non-dominated, spread-out subset without inflating validation cost. After scoring, a batch-relative step prunes consistently weak candidates, sparing later rounds from re-exploring them (Appendix~\ref{app:filter}).

\subsection{Selector}
\label{sec:selector}

Selection by a fixed scalar reward exhibits \emph{scalarization collapse}. Instead, the selector retains a population that is non-dominated and spread out, so the $k$ prompts carried to the next round span diverse accuracy-cost trade-offs. We use the non-dominated sorting of NSGA-II~\citep{deb2002fast}, which partitions candidates by their $(\mu_i, c_i)$ pairs into successive Pareto fronts $F_1, F_2, \dots$; these fronts are admitted whole, best front first, until the population budget is reached (Appendix~\ref{app:primer_nsga}). When admitting the next front would overflow the budget, the selector chooses among the candidates on that front (the \emph{boundary front} $F_j$) by a \emph{maximin spread rule}~\citep{gonzalez1985clustering}: starting from the two objective extremes of $F_j$ (highest $\tilde\mu$ and lowest $\tilde c$), it iteratively admits the candidate that maximizes the minimum Euclidean distance to the already-admitted candidates of $F_j$ in normalized $(\tilde\mu_i, \tilde c_i)$ space, until $k$ are filled, spreading the boundary-front picks across the accuracy-cost plane. Algorithm~\ref{alg:selector} and Figure~\ref{fig:selector} illustrate the selection process.

\paragraph{The Full Loop.}
Figure~\ref{fig:framework} shows the round-by-round loop. After $R$ rounds \CRAFT returns $\mathcal{P}_R$, a finite approximation to the accuracy-cost Pareto front; Table~\ref{tab:method_notation} in Appendix~\ref{app:notation} collects all symbols. Two choices set \CRAFT apart from prior multi-objective prompt optimization: candidates come from complementary accuracy-oriented and cost-oriented generators rather than from one mutation source, and the front-relative gap decides \emph{which} candidates receive target-LLM validation in each round rather than allocating those validations by a scalar score over the candidate pool.

\section{Experiments and Evaluation}
\label{sec:experiments}

\subsection{Datasets}
\label{sec:datasets}
We evaluate \CRAFT on six English classification and reasoning datasets: BeaverTails \citep{ji2023beavertails}, GoEmotions \citep{demszky-etal-2020-goemotions}, and four BIG-Bench Hard tasks (DisambiguationQA, Causal Judgement, Formal Fallacies, Salient Translation) \citep{suzgun-etal-2023-challenging}. We use the dataset abbreviations BT, GoE, DQA, CJ, FF, and ST throughout. Initial prompts for each dataset follow the public release of \citet{kumar-etal-2025-sculpt} so that all methods start from the same point. Full per-dataset validation/test sizes, task-score definitions, and task details are in Appendix~\ref{app:datasets}.

\subsection{Setup}
\CRAFT uses LLMs in two roles: an \emph{optimizer LLM} powers the critic and actor inside the refiner and condenser (i.e., it proposes prompt edits), and a \emph{target LLM} executes candidate prompts on validation/test examples to score them. Main comparisons, ablations, and the population-size sweep use OpenAI GPT-5 \citep{singh2025openai} as both optimizer and target; the cross-LLM study fixes the target to GPT-5.4-mini and varies only the optimizer. The optimizer LLM uses temperature 0.7 to diversify generated candidates; the target LLM scores with temperature 0. The main method comparison averages two seeds; the ablation, population-size, and cross-LLM studies report a single seed.

\begin{algorithm}[!t]
\footnotesize
\caption{\CRAFT Selection Procedure}
\label{alg:selector}
\begin{algorithmic}[1]
\Require Candidates $\{(\mu_i, c_i)\}_{i=1}^{2k}$, population size $k$
\State Compute $(\tilde{\mu}_i, \tilde{c}_i)$ by min-max normalizing $\{\mu_i\}, \{c_i\}$
\State $\{F_1, F_2, \dots\} \gets \textsc{FastNonDominatedSort}(\{(\mu_i, c_i)\})$
\State $\mathcal{P}_t \gets \emptyset$;\quad $j \gets 1$
\While{$|\mathcal{P}_t| + |F_j| \le k$}
 \State $\mathcal{P}_t \gets \mathcal{P}_t \cup F_j$;\quad $j \gets j + 1$
\EndWhile
\If{$|\mathcal{P}_t| < k$} \Comment{partial fill from $F_j$ by maximin spread}
 \State $r \gets k - |\mathcal{P}_t|$ \Comment{still to fill}
 \State $S \gets$ up to $\min(2, r)$ objective-extreme candidates of $F_j$
 \While{$|S| < r$ \textbf{and} $F_j\setminus S \neq \emptyset$}
 \State $i^{*} \gets \argmax\limits_{i \in F_j\setminus S}\ \min\limits_{s \in S}\ \|(\tilde{\mu}_i,\tilde{c}_i) - (\tilde{\mu}_s,\tilde{c}_s)\|_2$
 \State $S \gets S \cup \{i^{*}\}$
 \EndWhile
 \State $\mathcal{P}_t \gets \mathcal{P}_t \cup S$
\EndIf
\State \Return $\mathcal{P}_t$
\end{algorithmic}
\end{algorithm}

We set $R=8$ for all method comparisons; diagnostic figures use extended runs where noted, and \S\ref{sec:round_budget} shows best score and hypervolume flatten near $R=8$. Per prompt in the current population, \CRAFT's refiner produces up to \(n_R = 8\) refined variants and its condenser \(n_C = 4\) condensed variants; with population size \(k = 4\), full \CRAFT generates up to \(k(n_R{+}n_C)\) new candidates per round and retains \(k\) prompts. $B = 2k$ candidates are validated per round on the round's subset $\mathcal{V}_{\pi(t)}$, costing $B \cdot |\mathcal{V}_{\pi(t)}|$ LLM calls. Across comparison groups we keep $R$, $k$, the validation-subset schedule, and $B$ fixed; generation modules and score objectives vary by method as defined below. Target-validation compute is thus comparable; \CRAFT's advantage shows up as wider front coverage and lower deployment-time prompt-token cost.

\begin{figure*}[!t]
\centering
\includegraphics[width=\linewidth]{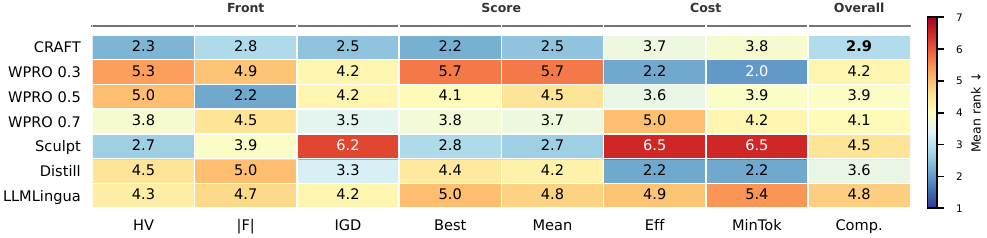}
\caption{Mean rank across the six benchmarks at $R{=}8$ (lower is better). Columns report ranks rather than raw metric values; the left columns group per-metric ranks into front, score, and cost aspects, and the rightmost \textbf{Overall} column is the composite rank with the three aspects weighted equally, lowest for \CRAFT.}
\label{fig:per_metric_rank}
\end{figure*}

\subsection{Variants and Baselines}
\label{sec:variants}

We compare \CRAFT with two baseline families and two component ablations.

\paragraph{Weighted-Sum Scalarization (\WPRO).}
CAPO scalarizes accuracy against a length-penalty term using a hyperparameter $\gamma$ fixed before search. Because accuracy and token-length ranges differ across datasets, one fixed $\gamma$ does not transfer; we address this by min-max-normalizing both axes over the current candidate set and taking a weighted sum, which we call \WPRO. It scores each candidate $P_i$ by $\mathrm{Score}(P_i) = w_p\hat{p}_i + w_c\hat{c}_i$, where $\hat{p}_i$ and $\hat{c}_i$ are normalized accuracy and token reduction (both larger-is-better; formulas in Appendix~\ref{app:modules}). In \WPRO, scalarization is used both when allocating validation and when retaining the population. For acquisition, we replace Pareto-gap UCB with a weighted-sum UCB score over normalized optimistic accuracy and token reduction; for retention, we sort validated candidates by the corresponding weighted score over empirical mean accuracy and token reduction and keep the top $k$. The refiner, condenser, validation subsets, and per-round validation batch size match \CRAFT. Because the weight is still fixed before search, \WPRO often recovers only a narrow region of the front (Figure~\ref{fig:motivation}), our empirical instance of \emph{scalarization collapse}. We sweep $w_p \in \{0.3, 0.5, 0.7\}$ with $w_c = 1 - w_p$ to probe cost-leaning, balanced, and score-leaning regimes of the trade-off.

\paragraph{Single-Axis Baselines.}
Single-axis baselines run inside \CRAFT's loop (same validation subsets and per-round budget) but use one generator family at a time and score-only acquisition/retention~\citep{pryzant-etal-2023-automatic,kumar-etal-2025-sculpt}, with no cost objective. Validation is allocated by score-only UCB, and the top $k$ candidates by empirical mean task score are retained. \SCULPT uses the accuracy-oriented refiner alone; \Distill and \LLMLingua use cost-oriented compression generators alone. Full setup details are in Appendix~\ref{app:modules}.

\paragraph{Component Ablations of \CRAFT.}
We isolate each generation module with two ablations: \CRAFT-R keeps only the refiner (\SCULPT, no condenser), and \CRAFT-C keeps only the condenser (\Distill, no refiner). To probe how much the refiner matters, we also run two alternate-refiner variants, {\CRAFT}$_{\text{OPRO}}$ and {\CRAFT}$_{\text{Evo}}$, which replace the default \SCULPT refiner with \OPRO and \Evo respectively while keeping the condenser, acquisition, and selector unchanged. Unlike the single-axis baselines, these ablations keep \CRAFT's Pareto-gap acquisition and NSGA-II selector.

\subsection{Evaluation Metrics}
\label{sec:metrics}
At the default $R=8$ snapshot, we report seven metrics covering score, cost, and front coverage. A prompt is \emph{feasible} if its test score is at least $0.95\times$ the initial-prompt test score; cost-leaning metrics use this filter to enforce the deployment floor below which cost savings are rarely worth the accuracy drop. This feasibility check is a reporting filter applied when computing metrics; it is not the in-loop pruning rule used during optimization (Appendix~\ref{app:filter}). The seven metrics are: \textbf{Best score} (max task score in the retained population), \textbf{Mean retained-population score} (\textbf{Mean} in tables and figures), \textbf{Min feasible tokens} (Tok), \textbf{Peak efficiency} (Eff = $E_{\max}$; the score-per-token ratio reported in percent-score-points per 100 tokens, i.e., $E_{\max} = 100\cdot p_{i^{*}} / c_{i^{*}}$ where $i^{*}=\arg\max_i p_i/c_i$), \textbf{Hypervolume} HV \citep{zitzler1999multiobjective} relative to the initial prompt, \textbf{Inverted Generational Distance} IGD \citep{bosman2003igd} against the method-union front, and \textbf{Front size} $|\mathcal{F}|$. Full per-metric definitions, the HV reference choice, and the method-union IGD justification are in Appendix~\ref{app:metrics_full}. The \emph{composite rank} aggregates per-metric ranks across three aspects of equal weight: front quality $\mathcal{A}_{\mathrm{q}}=\{\text{HV},|\mathcal{F}|,\text{IGD}\}$, score $\mathcal{A}_{\mathrm{s}}=\{\text{best},\text{mean}\}$, and cost $\mathcal{A}_{\mathrm{c}}=\{E_{\max},\text{tokens}\}$.
\vspace{-0.5em}
\begin{equation}
\mathrm{Composite}(m) = \frac{1}{3|D|}\sum_{d \in D}\sum_{\mathcal{A}} r_{\mathcal{A}}(m, d),
\end{equation}
\vspace{-1em}

\noindent where $r_{\mathcal{A}}(m, d)$ is the mean per-metric rank of method $m$ on dataset $d$ within aspect $\mathcal{A}$.

\section{Results}
\label{sec:results}

\paragraph{Method Comparison.}
\label{sec:wpro}
Figure~\ref{fig:per_metric_rank} reports each method's mean rank on every metric across the six datasets and aggregates them into the composite rank (rightmost column). Figure~\ref{fig:composite_rank_heatmap} gives per-dataset composite ranks; Table~\ref{tab:full_results} reports the absolute per-dataset values (Appendix~\ref{app:full_results}). The baselines exhibit \emph{scalarization collapse}. Single-axis baselines rank near the top of the axis they emphasize (\SCULPT 2nd on score, \Distill 2nd on cost) but trail on the others, and \LLMLingua fares worst because many of its compressed prompts fall below the feasibility threshold and lose even on cost. Weighted-sum baselines collapse along the trade-off weight: each \WPRO variant leads the axis its weight favors but trails the other, and even the balanced {\WPRO}$_{0.5}$ only reaches 3rd on the composite rank. \textbf{\CRAFT ranks 1st on the overall composite rank.} It does not win every metric, placing 3rd on cost behind baselines with cost-oriented generators, but maintains coverage of both high-accuracy and low-token regions: Figure~\ref{fig:motivation} shows on BT that each baseline confines its retained set to one region while \CRAFT's retained set covers both ends. Per-dataset rankings (Figure~\ref{fig:composite_rank_heatmap}) further show this lead reflects robustness across the three aspects rather than dominance on any individual dataset.

\paragraph{Front Evolution.}
\label{sec:dynamics}

\begin{figure}[!t]
\centering
\includegraphics[width=\linewidth]{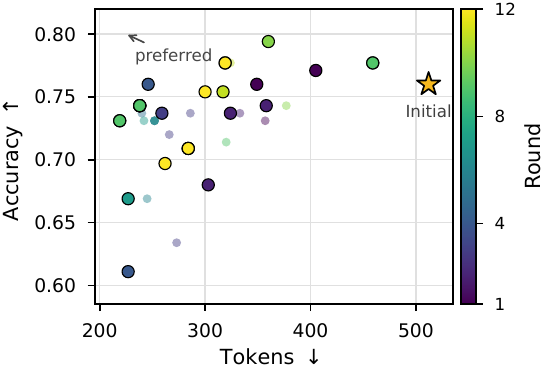}
\caption{\CRAFT's retained set on DQA across rounds. Non-dominated points have darker outlines; dominated ones are smaller; the gold star marks the initial prompt.}
\label{fig:combined_pareto}
\end{figure}

Figure~\ref{fig:combined_pareto} visualizes the prompts \CRAFT selects across an extended 12-round DQA run, colored by round. The selected prompts move toward the upper-left region of high accuracy and low cost: later rounds (lighter) reach both higher accuracy and lower cost than the initial prompt (gold star), and each round's non-dominated candidates spread along the trade-off rather than piling at a single corner. Per-metric trajectories across all six benchmarks (Appendix~\ref{app:dynamics}) improve overall but non-monotonically, as refinement and condensation alternately add and remove structure.

\begin{table}[H]
\centering
\footnotesize
\setlength{\tabcolsep}{3pt}
\resizebox{\columnwidth}{!}{%
\begin{tabular}{llrrrrrr}
\toprule
\textbf{Data} & \textbf{Setting} & \textbf{Best $\uparrow$} & \textbf{Mean $\uparrow$} & \textbf{Tok $\downarrow$} & \textbf{Eff $\uparrow$} & \textbf{HV $\uparrow$} & \textbf{$|\mathcal{F}|$ $\uparrow$} \\
\midrule
\multirow{5}{*}{BT} & \CRAFT-R & \textbf{63.0} & 61.5 & 1{,}013 & 6.2 & 62k & \textbf{3} \\
 & \CRAFT-C & 53.0 & 52.2 & \textbf{279} & \textbf{19.0} & 52k & 1 \\
 & \CRAFT & \textbf{63.0} & \textbf{61.7} & 512 & 12.3 & \textbf{63k} & \textbf{3} \\
 & {\CRAFT}$_{\text{OPRO}}$ & 59.0 & 57.3 & 590 & 9.5 & 57k & 2 \\
 & {\CRAFT}$_{\text{Evo}}$ & 54.0 & 53.5 & 326 & 16.3 & 53k & 1 \\
\midrule
\multirow{5}{*}{GoE} & \CRAFT-R & \textbf{41.0} & \textbf{40.5} & 2{,}130 & 1.9 & \textbf{39k} & 2 \\
 & \CRAFT-C & 32.0 & 31.0 & 400 & 8.0 & 31k & 2 \\
 & \CRAFT & 37.0 & 35.0 & 563 & 6.6 & 36k & 2 \\
 & {\CRAFT}$_{\text{OPRO}}$ & 38.0 & 36.8 & \textbf{318} & \textbf{11.3} & 38k & \textbf{3} \\
 & {\CRAFT}$_{\text{Evo}}$ & 36.0 & 34.8 & 455 & 7.3 & 35k & 2 \\
\bottomrule
\end{tabular}}
\caption{Ablation of \CRAFT on BT and GoE. \CRAFT-R retains the refiner only; \CRAFT-C retains the condenser only. {\CRAFT}$_{\text{OPRO}}$ and {\CRAFT}$_{\text{Evo}}$ replace the default \SCULPT refiner with \OPRO and \Evo. Front size counts feasible prompts only.}
\label{tab:ablation}
\end{table}

\paragraph{Ablation Study.}
\label{sec:ablations}
To isolate the contribution of each generation module, we run two \CRAFT variants: \CRAFT-R retains only the refiner; \CRAFT-C retains only the condenser. \CRAFT-R matches the full \CRAFT on accuracy and front size, but its prompts grow to $1{,}013$ tokens on BT and $2{,}130$ on GoE, far longer than the full \CRAFT ($512$ and $563$), because the refiner has no cost counterweight. \CRAFT-C produces the shortest prompts but reduces the best score by $10$ points on BT and $5$ on GoE, and on BT its front collapses to a single feasible prompt as aggressive compression pushes most variants below the feasibility threshold. Neither module suffices alone, since the refiner recovers accuracy but inflates length while the condenser shortens prompts but cannot recover score; only the full \CRAFT balances both axes.

\paragraph{Impact of Refiner.}
\label{sec:refiner_impact}
Table~\ref{tab:ablation} also reports two \CRAFT variants that change only the refiner, replacing \SCULPT with \OPRO or \Evo while keeping the condenser, acquisition, and selector fixed. {\CRAFT}$_{\text{OPRO}}$ matches or surpasses \CRAFT on GoE while trailing on BT, plausibly because \OPRO uses prior prompts and their validation scores to guide refinement. {\CRAFT}$_{\text{Evo}}$ improves over the initial prompt on GoE but lags on BT. Together they show that \CRAFT accommodates alternative refiners without retuning the loop, although the choice of refiner affects which dataset gains most.

\begin{table}[H]
\centering
\footnotesize
\setlength{\tabcolsep}{3pt}
\begin{tabular}{lcccccc}
\toprule
\textbf{Optimizer} & \textbf{Best $\uparrow$} & \textbf{Mean $\uparrow$} & \textbf{Tok $\downarrow$} & \textbf{Eff $\uparrow$} & \textbf{HV $\uparrow$} & \textbf{$|\mathcal{F}|$ $\uparrow$} \\
\midrule
Initial & 36.5 & 36.5 & 794 & 4.6 & n/a & 1 \\
{\CRAFT}$_{5.5}$ & \textbf{53.0} & 38.8 & 546 & 9.7 & \textbf{52k} & \textbf{4} \\
{\CRAFT}$_{5.4}$ & 51.0 & 35.5 & 562 & 9.1 & 50k & 3 \\
{\CRAFT}$_{\text{mini}}$ & 48.0 & 34.8 & 573 & 8.4 & 46k & 3 \\
{\CRAFT}$_{\text{DS}}$ & 51.0 & 36.0 & \textbf{502} & \textbf{10.2} & 51k & \textbf{4} \\
{\CRAFT}$_{\text{Kimi}}$ & 52.0 & \textbf{44.8} & 690 & 7.5 & 50k & 3 \\
\bottomrule
\end{tabular}
\caption{Cross-LLM transfer on GoE with GPT-5.4-mini as the fixed target. Optimizer subscripts: $5.5$ = GPT-5.5, $5.4$ = GPT-5.4, mini = GPT-5.4-mini, DS = DeepSeek-V4 Flash, Kimi = Kimi-K2.5.}
\label{tab:crossllm_goe}
\end{table}

\paragraph{Cross-LLM Transfer.}
\label{sec:crossllm}
\CRAFT's search operators depend on the optimizer LLM; the target LLM is held fixed while prompts are optimized. The main results use GPT-5 as both optimizer and target. To study the impact of the optimizer LLM, we fix the target at GPT-5.4-mini and run \CRAFT with five optimizers: GPT-5.5, GPT-5.4, GPT-5.4-mini, DeepSeek-V4 Flash \citep{deepseekai2026deepseekv4}, and Kimi-K2.5 \citep{team2026kimi}. On GoE (Table~\ref{tab:crossllm_goe}), every tested optimizer expands the front from the initial prompt, improving best score, peak efficiency, and hypervolume while reducing prompt-token cost, so \CRAFT's gain is not tied to matching optimizer and target, and works across optimizer LLM families. Optimizer choice shifts which metric leads: GPT-5.5 leads on best score and hypervolume, DeepSeek-V4 Flash on prompt-token cost and peak efficiency, and Kimi-K2.5 on mean retained-population score, so \CRAFT's front gains carry across optimizer families even when no one optimizer dominates every axis.

\begin{figure}[!t]
\centering
\includegraphics[width=\linewidth]{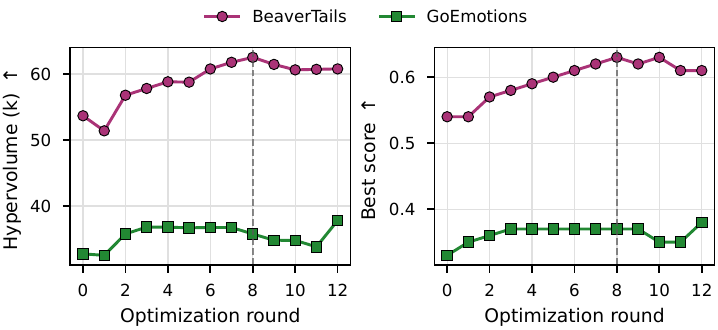}
\caption{\CRAFT hypervolume ($\uparrow$) and best score ($\uparrow$) over rounds $0$ to $12$ on BT and GoE.}
\label{fig:round_budget}
\end{figure}

\paragraph{Impact of Round Budget.}
\label{sec:round_budget}

To justify the $R=8$ default, we extend \CRAFT to $R{=}12$ on BT and GoE and track hypervolume and best score round by round. In Figure~\ref{fig:round_budget}, both metrics rise sharply early and flatten near $R=8$; the four extra rounds bring oscillation rather than steady gains. Reporting at $R=8$ captures most progress while keeping cost comparable across baselines.

\begin{table}[H]
\centering
\footnotesize
\setlength{\tabcolsep}{3pt}
\renewcommand{\arraystretch}{0.95}
\begin{tabular}{lrrrrrr}
\toprule
\textbf{Setting} & \textbf{Best $\uparrow$} & \textbf{Mean $\uparrow$} & \textbf{Tok $\downarrow$} & \textbf{Eff $\uparrow$} & \textbf{HV $\uparrow$} & \textbf{$|\mathcal{F}|$ $\uparrow$} \\
\midrule
BT, $k{=}2$ & 58.0 & 57.0 & 388 & 14.9 & 58k & 2 \\
BT, $k{=}4$ & \textbf{63.0} & \textbf{61.7} & 512 & 12.3 & \textbf{63k} & 3 \\
BT, $k{=}6$ & 61.0 & 58.3 & 349 & 17.5 & 61k & \textbf{4} \\
BT, $k{=}8$ & 54.0 & 52.1 & \textbf{244} & \textbf{22.1} & 52k & 2 \\
\midrule
GoE, $k{=}2$ & 37.0 & 34.5 & 602 & 6.1 & 37k & 2 \\
GoE, $k{=}4$ & 37.0 & 35.0 & 563 & 6.6 & 36k & 2 \\
GoE, $k{=}6$ & 40.0 & 35.3 & 860 & 4.7 & \textbf{40k} & 3 \\
GoE, $k{=}8$ & \textbf{42.0} & \textbf{36.3} & \textbf{435} & \textbf{7.4} & 37k & \textbf{4} \\
\bottomrule
\end{tabular}
\caption{Population-size ($k$) sweep for \CRAFT.}
\label{tab:kstudy}
\end{table}

\paragraph{Impact of Population Size.}
\label{sec:ksweep}

The retained-population size $k$ sets how many prompts the optimizer keeps each round. We use $k=4$ as the default to leave room for high-accuracy, short, balanced, and intermediate prompts in the retained population. Table~\ref{tab:kstudy} sweeps $k\in\{2,4,6,8\}$: $k=2$ is too small to cover both axes, while larger $k$ trades peak accuracy for broader or cheaper fronts. The two datasets respond differently: BT peaks at $k=4$ on best score and hypervolume but reaches its lowest token count and highest efficiency at $k=8$, while GoE improves nearly monotonically, winning best score, efficiency, and front size at $k=8$. The best $k$ therefore depends on whether score, cost, or front coverage matters most; $k=4$ is a balanced default across all six benchmarks.

\section{Conclusion}

In summary, \CRAFT demonstrates that cost-aware prompt optimization is more reliably addressed as Pareto-front search than as scalarized optimization. In our experiments, accuracy-only, cost-only, and weighted-sum baselines tend to concentrate in one region of the accuracy-cost front. \CRAFT does not win every metric, yet it remains competitive on each of front quality, task score, and cost and produces the strongest overall trade-offs, a balance that is preserved when the optimizer LLM is varied. The ablations show that the refiner and the condenser play complementary roles, since removing either restricts the candidate set available to the acquisition function, and only their combination provides the diversity needed to extend the front. The loop itself is not tied to \SCULPT or to any fixed implementation of its components, so alternative refiners or condensers can be plugged into the same front-aware procedure. \CRAFT is therefore most useful when the accuracy-cost trade-off is unknown a priori or varies across deployments.

\newpage
\section{Limitations}
\label{sec:limitations}

\CRAFT has several limitations. First, our evaluation is restricted to English classification and short-answer reasoning benchmarks. As a result, we cannot yet determine whether \CRAFT generalizes to more open-ended generation settings, such as summarization, code generation, dialogue, or multilingual prompting. Second, several aspects of our experimental setup remain fixed. While we vary the optimizer LLM (\S\ref{sec:crossllm}), the target model is kept within the OpenAI family. Evaluating \CRAFT on non-OpenAI targets, such as Claude or Gemini, as well as open-source model families including Llama, Qwen, and Mistral, remains an important direction for future work. Similarly, we do not study smaller target models in the 1B--7B parameter range.

Third, our main experiments use one default refiner and one default condenser, both powered by the same optimizer LLM. We partially examine the role of the refiner by replacing \SCULPT with \OPRO and \Evo (\S\ref{sec:ablations}). On the compression side, \LLMLingua-2 serves as a token-level alternative to \Distill; however, its relatively weak performance led us not to explore additional compressors within \CRAFT. Future work could investigate history-aware compressors inspired by \OPRO, evolutionary compressors based on \Evo, or alternative generator LLMs for both modules. Finally, \CRAFT assumes a minimally task-relevant initial prompt. Its behavior under very weak, underspecified, or adversarial initial prompts remains unexplored.

\bibliography{references}

\appendix

\section{Method Details}

\subsection{Notation}
\label{app:notation}

Table~\ref{tab:method_notation} lists the symbols used in Section~\ref{sec:framework}.

\begin{table}[H]
\small
\centering
\setlength{\tabcolsep}{3pt}
\begin{tabular}{p{1.25cm} p{4.95cm}}
\toprule
\textbf{Symbol} & \textbf{Definition} \\
\midrule
$R$ & Number of \CRAFT optimization rounds; main experiments use $R = 8$. \\
$T$ & Number of validation subsets; conceptually distinct from the optimization-round count $R$, set to $8$ by default. \\
$K_{\mathrm{cluster}}$ & Number of $k$-means clusters used to build validation subsets; $K_{\mathrm{cluster}} = T$. \\
$k$ & Population size retained by the selector after each round. \\
$n_R$ & Maximum number of refined candidates generated per round. \\
$n_C$ & Maximum number of condensed candidates generated per round. \\
$B$ & UCB-allocated validation batch size per round; default $B = 2k$. \\
$\mu_{i,t}$ & Running mean validation score for prompt $P_i$ through round $t$; we drop the $t$ subscript when context fixes the round. \\
$\sigma_{i,t}$ & Empirical standard deviation over observed validation scores; conservative $\sigma=1$ floor when $|\mathcal{S}_{i,t}|\le 1$. \\
$c_i$ & Prompt-token count of $P_i$. \\
$p^+_{i,t}$ & Optimistic score $p^+_{i,t}=\mu_{i,t}+\beta_t\sigma_{i,t}$ used for acquisition. \\
$\beta_t$ & Linear exploration weight $0.5+0.5\max(0,1-t/R)$, decaying over optimization rounds. \\
$B_{\mathrm{round}}$ & Per-round LLM-call budget: $k(n_R + n_C)$ generation calls plus $B\cdot|\mathcal{V}_{\pi(t)}|$ validation calls. \\
$\mathcal{P}_t$ & Retained population of size $k$ at round $t$. \\
$\mathcal{Q}_t$ & Candidate set at round $t$ (refined $\cup$ condensed $\cup$ retained). \\
$\mathcal{V}_j$ & The $j$-th validation subset, $j\in\{1,\dots,T\}$; optimization round $t$ validates on $\mathcal{V}_{\pi(t)}$ with $\pi(t)=((t-1)\bmod T)+1$. \\
$\mathcal{S}_{i,t}$ & Observation set for $P_i$: the validation examples on which it has been scored through round $t$. \\
$\mathcal{A}_t$ & Archive of all candidates evaluated through round $t$. \\
$\mathcal{F}^+_t$ & Optimistic Pareto front in $(\tilde{p}^+, \tilde{c})$ space at round $t$. \\
$\mathcal{E}_t$ & Evaluation batch selected at round $t$, $|\mathcal{E}_t|\le B$. \\
$\mathcal{C},\,\mathcal{C}^{*}$ & Generic candidate set and its Pareto-optimal subset (Section~\ref{sec:problem}). \\
\bottomrule
\end{tabular}
\caption{Notation used in the \CRAFT method section. We use subscripted forms ($\mu_i$, $\sigma_i$, $c_i$, $p_i$) throughout, adding the round subscript $t$ (e.g.\ $\mu_{i,t}$, $\sigma_{i,t}$, $p^{+}_{i,t}$) when a per-round running estimate is being defined.}
\label{tab:method_notation}
\end{table}

\subsection{Background Primer: NSGA-II and UCB}
\label{app:primer}

This appendix gives the methodological background for the two algorithmic components of \CRAFT that are reused from prior work: NSGA-II selection and UCB acquisition. Pareto-front and dominance are defined in Section~\ref{sec:problem} (Preliminaries); hypervolume and IGD are defined in Section~\ref{sec:metrics} and elaborated in Appendix~\ref{app:metrics_full}. Throughout this appendix we use \CRAFT's bi-objective setting: task score $p_i \in [0, 1]$ (maximize) and prompt-token cost $c_i \in \mathbb{Z}_{>0}$ (minimize), with the same per-candidate index convention as in the main paper.

\paragraph{Why a Weighted-Sum Cannot Replace Front Search.}
A weighted-sum scalarization $w\,\hat{p}_i + (1-w)\,\hat{c}_i$ (used by our \WPRO baselines) commits to a single tangent of the front, parameterized by $w$. It is provably unable to recover non-convex regions of the Pareto front for any choice of $w$~\citep{das1997closer}: scalarization is fundamentally weaker than direct front search when the trade-off shape is unknown a priori. This is the formal statement of the \emph{scalarization-collapse} failure mode named in Section~\ref{sec:intro} and motivates \CRAFT's front-aware acquisition.

\subsubsection{Front-aware selection: non-dominated sort and maximin spread}
\label{app:primer_nsga}

\CRAFT's selector reuses fast non-dominated sort from NSGA-II~\citep{deb2002fast} and pairs it with a maximin spread tie-breaker~\citep{gonzalez1985clustering} instead of the standard NSGA-II crowding-distance estimator. We describe both ingredients.

\paragraph{Fast Non-Dominated Sort.}
Given a candidate set of size $N$ with $K$ objectives, the sort partitions candidates into successive Pareto fronts $F_1, F_2, \dots$: $F_1$ contains candidates not dominated by any other, $F_2$ contains candidates not dominated by any candidate outside $F_1$, and so on. The procedure runs in $O(K N^2)$.

\paragraph{Maximin Spread within a Front.}
Let $\tilde{\mu}_i$, $\tilde{c}_i$ be min-max normalized estimates in $[0,1]$. From the boundary front $F_j$ we want to pick $m=k-|\mathcal{P}_t|$ candidates that are spread along the front. We use the classical farthest-first traversal~\citep{gonzalez1985clustering}: seed the selected set $S$ with the two extremes of $F_j$ (the candidates that maximize $\tilde{\mu}$ and minimize $\tilde{c}$ respectively); then while $|S|<m$, add $i^{*}=\argmax_{i\in F_j\setminus S}\min_{s\in S}\|(\tilde{\mu}_i,\tilde{c}_i)-(\tilde{\mu}_s,\tilde{c}_s)\|_2$. Each iteration picks the candidate whose closest neighbor in $S$ is the farthest away, which spreads $S$ as evenly as possible along $F_j$ in normalized objective space. The traversal is seeded from the boundary front $F_j$'s own extremes rather than from the already-admitted fronts $F_1,\dots,F_{j-1}$; this keeps boundary-front selection self-contained, at the cost of not explicitly spacing the newly admitted points against the previously retained ones.

\paragraph{Selector.}
To form the next-round population of size $k$, the selector traverses $F_1, F_2, \dots$ in order, admitting whole fronts until the next would overfill. From the boundary front it picks the remaining candidates by maximin spread, as stated in Algorithm~\ref{alg:selector} of the main paper.

\begin{figure}[!t]
\centering
\includegraphics[width=\linewidth]{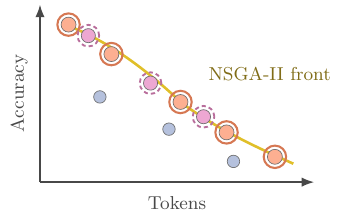}
\vspace{-1em}
\caption{\CRAFT selector. Filled circles are candidates in $(c,\,\mu)$ space; the yellow curve is the non-dominated front. Orange-ringed points are retained by the maximin spread rule; pink dashed-ring points are on the front but pruned because admitting them would shrink the minimum pairwise distance of the retained set; blue points are dominated. The rule jointly preserves non-dominance and even spread.}
\label{fig:selector}
\end{figure}

\subsubsection{Upper-Confidence-Bound (UCB) acquisition}
\label{app:primer_ucb}

UCB acquisition is a generic optimism-under-uncertainty rule from the bandit literature~\citep{auer2002finite,srinivas2009gaussian}. For candidate $P_i$ with mean estimate $\mu_i$ and uncertainty estimate $\sigma_i$, the UCB score is $\mathrm{UCB}_i = \mu_i + \beta\sigma_i$ with exploration weight $\beta > 0$. Candidates with high mean \emph{or} high uncertainty receive high UCB scores. As more observations accumulate, $\sigma_i$ shrinks and UCB approaches the empirical mean, so confidence intervals tighten with experience.

The original Gaussian-process UCB rule~\citep{srinivas2009gaussian} derives $\sigma$ from a GP posterior. \CRAFT does not fit a surrogate model over prompts and instead uses an empirical-SD variant. $\sigma_{i,t}$ is the standard deviation of $P_i$'s per-example task scores observed so far, with a conservative $\sigma = 1$ floor for candidates with $\le 1$ observation. The schedule $\beta_t = 0.5 + 0.5\max(0, 1 - t/R)$ starts near $1$ at early rounds and linearly decays to $0.5$ by the final optimization round $R$, never collapsing to pure exploitation.

\CRAFT's twist on UCB is multi-objective. Instead of one UCB ranking, the optimistic scores $p^+_{i,t} = \mu_{i,t} + \beta_t\sigma_{i,t}$ are paired with costs $c_i$ to form an optimistic front, and the acquisition prefers candidates closest to (or already on) that front. The full rule is in Algorithm~\ref{alg:pgucb}.

\paragraph{Mean and Uncertainty.}
For prompt $P_i$, let $\mathcal{S}_{i,t}$ be the validation examples on which $P_i$ has been evaluated up to round $t$, let $z=(x,y)$, and $s_i(z)=s(P_i,x,y)\in[0,1]$ the per-example score. The running mean and empirical standard deviation are
\begin{align*}
\mu_{i,t} &= \frac{1}{|\mathcal{S}_{i,t}|}\sum_{z\in\mathcal{S}_{i,t}} s_i(z), \\
\sigma_{i,t} &= \sqrt{\frac{\sum_{z\in\mathcal{S}_{i,t}}\big(s_i(z)-\mu_{i,t}\big)^2}{\max(|\mathcal{S}_{i,t}|-1,\,1)}}.
\end{align*}
With no observations \CRAFT sets $\mu_{i,t}=0$, $\sigma_{i,t}=1$; with one observation it keeps a conservative $\sigma=1$ floor.

\paragraph{Normalized Score-Cost Coordinates.}
Let $\mathcal{A}_t$ be the archive of candidates evaluated up to round $t$. With
\(p^+_{\min}=\min_{j\in\mathcal{A}_t} p^+_{j,t},\ p^+_{\max}=\max_{j\in\mathcal{A}_t} p^+_{j,t}\) and
\(c_{\min}=\min_{j\in\mathcal{A}_t} c_j,\ c_{\max}=\max_{j\in\mathcal{A}_t} c_j\), the normalized coordinates are
\begin{align*}
\tilde{p}_{i,t} &= \frac{p^+_{i,t}-p^+_{\min}}{p^+_{\max}-p^+_{\min}+\epsilon}, \\
\tilde{c}_i &= \frac{c_i-c_{\min}}{c_{\max}-c_{\min}+\epsilon},
\end{align*}
with $\epsilon$ a small constant to guard against constant archives. Higher $\tilde{p}$ and lower $\tilde{c}$ are preferred.

\paragraph{Complexity.}
Algorithm~\ref{alg:pgucb} runs in $O(|\mathcal{Q}_t| \log |\mathcal{Q}_t|)$ for two-objective front construction, $O(|\mathcal{Q}_t|\cdot|\mathcal{F}^+_t|)$ for per-candidate gap computation (bounded by $O(|\mathcal{Q}_t|^2)$ in the worst case), and $O(|\mathcal{Q}_t| \log |\mathcal{Q}_t|)$ for the final sort.

\paragraph{Pareto Gap.}
For candidate $i$ the gap to the optimistic front $\mathcal{F}^+_t$ is
\[
g_i = \min_{y\in\mathcal{F}^+_t}\!\big[\,\max(0,\tilde{p}_{y,t}\!-\!\tilde{p}_{i,t}) + \max(0,\tilde{c}_i\!-\!\tilde{c}_y)\,\big].
\]
The two $\max(0,\cdot)$ terms count score deficit (how much less $\tilde{p}_{i,t}$ is than a front point's $\tilde{p}_{y,t}$) and cost excess (how much more $\tilde{c}_i$ is than the same front point's $\tilde{c}_y$). The closest front point in this Manhattan sense determines $g_i$. Geometrically, $g_i=0$ iff $P_i$ is already non-dominated in the normalized space.

\subsection{\SCULPT and \Distill: Module Details}
\label{app:modules}

\begin{figure}[!t]
\centering
\includegraphics[width=\linewidth]{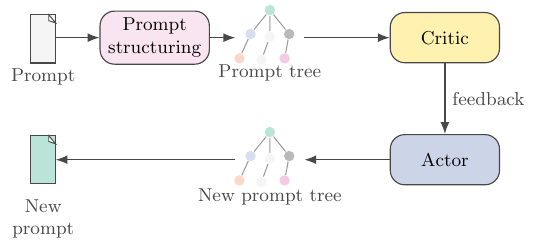}
\vspace{-1em}
\caption{Shared critic-actor scaffold for the default refiner (\SCULPT) and condenser (\Distill) of \CRAFT. The input prompt is parsed into a hierarchical tree; the critic emits feedback to the actor, which applies tree-level edits to produce a revised prompt.}
\label{fig:actor_critic}
\end{figure}

\CRAFT's two candidate generators, \SCULPT (refiner, Section~\ref{sec:refiner}) and \Distill (condenser, Section~\ref{sec:condenser}), share the critic-actor scaffold shown in Figure~\ref{fig:actor_critic}. They differ in the roles assigned to the critic and the actor and in the action sets used (full details below).

\paragraph{Prompt-Tree Representation.}
Both modules operate on a hierarchical \emph{prompt tree}: each node is a Heading (\texttt{H}), Paragraph (\texttt{P}), or List-item (\texttt{L}) with a level, sibling order, GUID, and token count. The tree is parsed once at the start of each module call. Nodes are addressed as \texttt{<NodeType>-<Level>-<Order>-<GUID>}; deleting a node deletes its subtree.

\paragraph{\SCULPT Refiner.}
We adapt \SCULPT \citep{kumar-etal-2025-sculpt} as the default refiner. The \emph{critic} inspects the parsed prompt tree (Figure~\ref{fig:actor_critic}) together with the previous round's evaluation results, and emits feedback in two channels: \emph{structural feedback} (reorganization suggestions over the tree such as promote, split, or merge sections) and \emph{error feedback} (issues inferred from per-example predictions vs.\ ground truth, such as misclassified examples and ambiguous instructions). The \emph{actor} consumes the redacted tree and the critic's feedback and outputs a justified sequence of node-level edits drawn from the action set in Table~\ref{tab:sculpt_actions}. Each call returns one revised prompt tree; the actor is called $n_R$ times per parent per round.

\begin{table}[H]
\small
\centering
\setlength{\tabcolsep}{3pt}
\begin{tabular}{p{1.8cm} p{4.7cm}}
\toprule
\textbf{Action} & \textbf{Description} \\
\midrule
\texttt{UPDATE}  & Rewrite a node to clarify instructions, narrow or broaden scope, or fix unclear phrasing. \\
\texttt{INSERT}  & Add a new node (definition, example, or edge-case rule) under a parent. \\
\texttt{DELETE}  & Remove a redundant, contradictory, or off-target node and its subtree. \\
\texttt{MERGE}   & Combine overlapping sibling nodes into one. \\
\texttt{REORDER} & Change sibling order to improve logical flow. \\
\bottomrule
\end{tabular}
\caption{Action types in \SCULPT for structure-aware refinement. Each action operates on the parsed prompt tree of Figure~\ref{fig:actor_critic}.}
\label{tab:sculpt_actions}
\end{table}

\paragraph{\Distill Condenser.}
\Distill reuses the same critic-actor scaffold for cost-oriented edits. The \emph{critic} receives the token-annotated prompt tree and a target compression percentage, and emits per-node feedback respecting four constraints: preserve essential task information, use scoped (node-local) actions, respect tree hierarchy (deleting a node deletes its subtree), and meet the target reduction. The \emph{actor} consumes the critic feedback and applies the action set in Table~\ref{tab:distill_actions} to produce a compressed tree whose token count meets the target.

\begin{table}[H]
\small
\centering
\setlength{\tabcolsep}{3pt}
\begin{tabular}{p{2cm} p{4.5cm}}
\toprule
\textbf{Action} & \textbf{Description} \\
\midrule
Update Node & Reword or simplify a single node. \\
Update Subtree & Simplify an entire subtree. \\
Delete Node & Remove redundant sections. \\
Merge Nodes & Combine overlapping sibling nodes. \\
\bottomrule
\end{tabular}
\caption{Action types in \Distill for structure-aware compression. Each action operates on the parsed prompt tree.}
\label{tab:distill_actions}
\end{table}

\paragraph{Compression Ratio.}
Per round, \Distill is called $n_C$ times with target compression ratios sampled from $[\rho_{\min}, \rho_{\max}]$, where the ratio is the fraction of prompt tokens to remove. In our experiments $\rho_{\min}=0.10$, $\rho_{\max}=0.20$, and $n_C=4$; the four sampled ratios are passed as inputs to the critic prompt, so the critic emits a per-node plan that targets each ratio. The actor enforces the target by adjusting the depth/breadth of deletions and merges.

\paragraph{\OPRO Refiner.}
The \OPRO refiner variant replaces the \SCULPT critic-actor pair with a single meta-prompt over the validated history of past prompts. Its inputs are (i) the current prompt $P_i$ and (ii) a history block of the top-$k$ validated past prompts sorted by validation score (best last). The optimizer LLM is asked to produce one new prompt that scores higher than every prompt in the history block. We issue the meta-prompt $n_R$ times per parent with seeded temperature variation and a per-call distinct-rewrite instruction, so the LLM produces $n_R$ different candidates rather than collapsing to a cached response. \OPRO operates on raw prompt text rather than the parsed tree.

\paragraph{\Evo Refiner.}
The \Evo variant replaces the \SCULPT critic-actor pair with a differential-evolution (DE) operator. Per parent, three validated prompts $P_1, P_2, P_3$ are sampled from the retained population and paired with the parent ($P_i$). The optimizer LLM is asked to (i) identify the differences between $P_1$ and $P_2$, (ii) mutate those differences, (iii) combine the mutation with $P_3$ to produce an intermediate prompt, and (iv) cross over the intermediate prompt with the parent to produce the final candidate. We issue the DE prompt $n_R$ times per parent with different sampled triples. Like \OPRO, \Evo operates on raw prompt text rather than the parsed tree.

\paragraph{Codebase References.}
Full Markdown prompt templates for \SCULPT (structural critic, error critic, actor), \Distill (critic, actor), \OPRO (meta-prompt), and \Evo (DE operator), the action-set implementations, and the prompt-tree parser are released in the public codebase.\footnote{\url{https://github.com/Sshanu/CRAFT/tree/main/src/data/prompt_enhancer/}}

\paragraph{Baseline Construction for \Distill and \LLMLingua.}
All single-axis baselines run inside the \CRAFT loop. Each round they generate $n_R + n_C$ candidates per prompt (matching \CRAFT's total per-round generation budget) from a single module, validate them with the score-only UCB acquisition rule of \citet{pryzant-etal-2023-automatic} and \citet{kumar-etal-2025-sculpt}, and retain the top $k$ candidates by empirical mean task score. The baselines differ only in the generator: the \SCULPT baseline uses the \SCULPT refiner; the \Distill baseline uses the \Distill condenser; the \LLMLingua baseline uses \LLMLingua-2. All three thus share validation subsets, per-round budget, acquisition rule, and retention rule with \CRAFT, but use one-axis generation and score-only ranking.

\paragraph{\WPRO Normalization.}
The \WPRO baseline scores a candidate $P_i$ by the weighted sum $\mathrm{Score}(P_i) = w_p\hat{p}_i + w_c\hat{c}_i$, where $\hat{p}_i$ and $\hat{c}_i$ are the min-max-normalized accuracy and token reduction over the current candidate set:
\begin{equation*}
\begin{aligned}
\hat{p}_i &= \frac{\mu_i - \min_j \mu_j}{\max_j \mu_j - \min_j \mu_j},\\[2pt]
\hat{c}_i &= \frac{\max_j c_j - c_i}{\max_j c_j - \min_j c_j}.
\end{aligned}
\end{equation*}
Both lie in $[0,1]$, with larger values denoting higher accuracy and greater token reduction respectively.

\subsection{Validation Subset Construction Detail}
\label{app:val_subsets}

This appendix expands the one-paragraph description of validation subsets in Section~\ref{sec:approximator}.

\paragraph{Why Partial Validation.}
A full-validation round would evaluate every selected candidate on all $N_{\mathrm{val}}$ examples in $\mathcal{D}_{\mathrm{val}}$. With $B$ candidates evaluated per round across $R$ rounds, the total validation-call count would be $R \cdot B \cdot N_{\mathrm{val}}$. Partial validation replaces $N_{\mathrm{val}}$ with $|\mathcal{V}_{\pi(t)}|$ per round, cutting cost by a factor of $N_{\mathrm{val}} / |\mathcal{V}_{\pi(t)}|$. In the main experiments, $T=8$, $R=8$, $B=2k=8$, and $N_{\mathrm{val}}$ is taken from Table~\ref{tab:dataset_stats}; this saves a factor of $T=8$ in validation calls.

\paragraph{Why Clustered Subsets.}
$\mathcal{D}_{\mathrm{val}}$ is partitioned into $T$ clusters by $k$-means~\citep{macqueen1967kmeans} in embedding space, using \texttt{all-MiniLM-L6-v2} sentence embeddings~\citep{reimers-gurevych-2019-sentence}. Each cluster becomes one validation subset $\mathcal{V}_j$, $j\in\{1,\dots,T\}$. Compared with random sampling, clustered subsets give each round semantically coherent examples; this avoids the variance from random sampling, where one round's subset can be substantially easier than another's by chance. The trade-off is that any per-subset evaluation depends on the cluster's idiosyncrasies; this is mitigated because the running mean $\mu_{i,t}$ accumulates evidence across the subsets where $P_i$ has been evaluated.

\paragraph{Embedding-Model Choice.}
\texttt{all-MiniLM-L6-v2} is a 22M-parameter sentence transformer trained for general semantic similarity. It is small enough that the embedding step is negligible compared to LLM calls and gives meaningful clusters for short classification and reasoning inputs. We did not sweep across embedding models; this is flagged as a limitation (Section~\ref{sec:limitations}).

\paragraph{Subset Rotation and Cumulative Observations.}
The validation subset used at optimization round $t$ is $\mathcal{V}_{\pi(t)}$ with $\pi(t)=((t-1) \bmod T) + 1$ (1-indexed); with $T=8$ and the main-experiment $R=8$, the evaluator visits each subset exactly once across the run, and diagnostic runs with $R>T$ revisit subsets cyclically. For a candidate $P_i$ that is retained across multiple rounds, the observation set $\mathcal{S}_{i,t}$ is the union of all $\mathcal{V}_{\pi(\tau)}$ on which $P_i$ has been evaluated through round $t$. The running mean $\mu_{i,t}$ averages over $\mathcal{S}_{i,t}$, and the empirical standard deviation $\sigma_{i,t}$ uses the same per-example scores. As a result, long-lived candidates accumulate evidence across multiple subsets, and $|\mathcal{S}_{i,t}|$ can exceed $|\mathcal{V}_{\pi(t)}|$.

\paragraph{Initial-Round Special Case.}
Before the first optimization round, the evaluator measures the initial prompt's full validation score $\mu_{0,0}$ on all of $\mathcal{D}_{\mathrm{val}}$ rather than on a single subset $\mathcal{V}_j$. This avoids using a single-cluster subset to set the reference for subsequent feasibility checks (the $0.95 \cdot p_{\mathrm{init}}$ threshold in Section~\ref{sec:metrics}).

\subsection{Batch-Relative Candidate Pruning}
\label{app:filter}

Between validation and selection, \CRAFT prunes candidates that are consistently weak relative to the current round's batch, so later rounds are not spent re-exploring them. The rule is batch-relative and uncertainty-aware, and is deliberately lenient: it is \emph{not} the strict $0.95\,p_{\mathrm{init}}$ feasibility filter used for reporting (Section~\ref{sec:metrics}), so moderately weak prompts that may recover in later rounds are retained.

For each validated candidate $P_i$ we form a lower-confidence-bound score $\ell_i = \mu_{i,t} - \beta_t\sigma_{i,t}$. A single strictness knob $f\in[0,1]$ (set to $f=0.7$ in all runs) maps to a quantile $q$ and a margin $\delta$:
\[
\begin{aligned}
q &= q_{\min} + (q_{\max}-q_{\min})\,f,\\
\delta &= \delta_{\max} - (\delta_{\max}-\delta_{\min})\,f,
\end{aligned}
\]
with $(q_{\min},q_{\max})=(0.70,0.98)$ and $(\delta_{\min},\delta_{\max})=(0.05,0.15)$; at $f=0.7$ this gives $q\approx0.90$ and $\delta\approx0.08$. The pruning threshold blends the batch $q$-quantile with a margin below the batch maximum,
\[
\tau = \tfrac{1}{2}\big(\mathrm{quantile}_q(\{\ell_i\}) + (\textstyle\max_i \ell_i - \delta)\big),
\]
and candidates with $\ell_i \ge \tau$ are kept. A floor of $\lceil 1.5\,k\rceil$ candidates is always retained: if fewer pass, the top-$\lceil 1.5\,k\rceil$ by $\ell_i$ are kept instead. Because $\tau$ is defined from the batch, the rule adapts to each round's score distribution rather than imposing an absolute cutoff.

\section{Experimental Setup}

\subsection{Dataset Statistics}
\label{app:datasets}

Table~\ref{tab:dataset_stats} reports the validation/test split sizes used in our experiments. For BeaverTails and GoEmotions we cap the public splits at 200 and 500 examples respectively to keep validation cost comparable across datasets. For the four BIG-Bench Hard tasks the public splits are smaller than this cap, so the cap is non-binding and the counts shown are the dataset's own validation/test sizes.

\begin{table}[H]
\centering
\small
\setlength{\tabcolsep}{3pt}
\begin{tabular}{llrrl}
\toprule
\textbf{Abbr.} & \textbf{Dataset} & \textbf{$N_{\mathrm{val}}$} & \textbf{$N_{\mathrm{test}}$} & \textbf{Task score} \\
\midrule
BT & BeaverTails & 200 & 500 & weighted-F1 \\
GoE & GoEmotions & 200 & 500 & micro-F1 \\
DQA & DisambiguationQA & 50 & 175 & exact acc. \\
CJ & Causal Judgement & 37 & 130 & macro-F1 \\
FF & Formal Fallacies & 50 & 175 & macro-F1 \\
ST & Salient Translation & 50 & 175 & macro-F1 \\
\bottomrule
\end{tabular}
\caption{Validation/test split sizes per dataset, with the task-score definition used in our experiments. Counts reflect the smaller of (i) the dataset's public split and (ii) the 200/500 cap we apply uniformly across datasets.}
\label{tab:dataset_stats}
\end{table}

\paragraph{BeaverTails (BT).} A multi-class safety classification dataset of human-LLM dialogues \citep{ji2023beavertails}; each prompt-response pair carries 14 binary harm-category labels (e.g., \emph{discrimination, hate speech}, \emph{drug abuse}, \emph{financial crime}). We score with weighted-F1 across the 14 labels.

\paragraph{GoEmotions (GoE).} A 27-emotion classification dataset of English Reddit comments \citep{demszky-etal-2020-goemotions} including categories like \emph{joy, sadness, gratitude, anger, neutral}. Multi-label; we score with micro-F1.

\paragraph{DisambiguationQA (DQA).} A BIG-Bench Hard task \citep{suzgun-etal-2023-challenging} in which the model resolves a pronoun in a short sentence to its correct antecedent (often a gender-disambiguation example). Single-answer multiple choice; scored with exact accuracy.

\paragraph{Causal Judgement (CJ).} A BIG-Bench Hard task \citep{suzgun-etal-2023-challenging} that asks the model to judge whether one event causally explains another in a short paragraph; binary answer scored with macro-F1.

\paragraph{Formal Fallacies (FF).} A BIG-Bench Hard task \citep{suzgun-etal-2023-challenging} that asks whether a short argument is a syntactically valid deductive inference; binary answer scored with macro-F1.

\paragraph{Salient Translation Error Detection (ST).} A BIG-Bench Hard task \citep{suzgun-etal-2023-challenging} that asks the model to identify which of six error categories (e.g., \emph{named-entity, numeric, modifiers-adjectives}) best describes a translation mistake; six-way classification scored with macro-F1.

\subsection{Evaluation Metrics: Full Definitions}
\label{app:metrics_full}

This appendix expands Section~\ref{sec:metrics}. All metrics are computed on the final retained population at the chosen round snapshot.

\paragraph{Feasibility Filter.}
A prompt is \emph{feasible} if its test score is at least $0.95\times$ the initial-prompt test score. Cost-leaning metrics (min feasible tokens, peak efficiency) are computed only over the feasible subset to prevent rewarding degenerate prompts that are short only because they fail the task.

\paragraph{Per-Metric Definitions.}
\begin{itemize}
\item \textbf{Best score}: $\max_i p_i$ across the $k$ retained candidates.
\item \textbf{Mean retained-population score} (\textbf{Mean} in tables and figures): $(1/k)\sum_{i \in \mathcal{P}_t} p_i$ across the retained set $\mathcal{P}_t$.
\item \textbf{Min feasible tokens}: $\min_i c_i$ across feasible candidates only.
\item \textbf{Peak efficiency} (Eff = $E_{\max}$): $100\cdot \max_i p_i/c_i$ across feasible candidates, reported in percent-score-points per 100 tokens (matching the main-text definition in \S\ref{sec:metrics}).
\item \textbf{Hypervolume} (HV) \citep{zitzler1999multiobjective}: the area of the $(p, c)$-plane dominated by the retained front, measured against a fixed worst-corner reference point at zero accuracy and a deliberately large token count ($r = (0,\,c_{\mathrm{ref}})$ with $c_{\mathrm{ref}}=100000$). This reference acts as the origin of the dominated-volume calculation: higher tokens (worse cost) sit at the origin, so retained prompts at lower cost and higher accuracy contribute more area. The reference is chosen far beyond any prompt observed in our runs so every retained candidate contributes positively. Reported HV is the raw value normalized to the initial-prompt corner $(p_{\mathrm{init}}, c_{\mathrm{init}})$, making HV comparable within a dataset across methods; we do \emph{not} compare HV across datasets because the references differ.
\item \textbf{Inverted Generational Distance} (IGD) \citep{bosman2003igd}: $\mathrm{IGD}(F; F^{*}) = (1/|F^{*}|)\sum_{q\in F^{*}}\min_{p\in F} d(q,p)$, where $d$ is Euclidean distance in min-max normalized $(\tilde{p}, \tilde{c})$ coordinates. Classical IGD uses an externally known true front; for each dataset we substitute the union of all methods' feasible fronts as $F^{*}$, since no externally validated Pareto front is available for prompt search. IGD penalizes fronts that leave regions of $F^{*}$ uncovered, complementing HV which rewards dominated area.
\item \textbf{Front size} ($|\mathcal{F}|$): the cardinality of the final non-dominated set. Reported for continuity with prior multi-objective evolutionary algorithm (MOEA) work, but it is bounded by $k$.
\end{itemize}

\paragraph{Composite Rank.}
For each metric $x$ and dataset $d$ we rank the compared methods from 1 to $|M|$ (1 is best). The composite first averages ranks within each aspect group (front quality $\mathcal{A}_{\mathrm{q}}$, score $\mathcal{A}_{\mathrm{s}}$, cost $\mathcal{A}_{\mathrm{c}}$), then averages across the three groups and across datasets. As a worked example, \CRAFT's per-metric mean ranks, read from Figure~\ref{fig:per_metric_rank}, are HV 2.33, $|\mathcal{F}|$ 2.83, IGD 2.50 (so $r_{\mathcal{A}_{\mathrm{q}}} = 2.55$); best 2.25, mean 2.50 (so $r_{\mathcal{A}_{\mathrm{s}}} = 2.375$); peak efficiency 3.67, min feasible tokens 3.83 (so $r_{\mathcal{A}_{\mathrm{c}}} = 3.75$). The composite is the mean across aspects: $(2.55 + 2.375 + 3.75)/3 = 2.89$.

\section{Additional Results}

\subsection{Full Per-Dataset Results}
\label{app:full_results}

Table~\ref{tab:full_results} reports the full per-dataset metrics that back the rank-based comparisons in \S\ref{sec:results}.

\begin{table*}[t]
\centering
\scriptsize
\setlength{\tabcolsep}{2.6pt}
\renewcommand{\arraystretch}{0.95}
\begin{minipage}[t]{0.49\textwidth}\centering
\begin{tabular}{lrrrrrrr}
\toprule
\multicolumn{8}{c}{\textbf{BT}} \\
\midrule
\textbf{Method} & \textbf{Best $\uparrow$} & \textbf{Mean $\uparrow$} & \textbf{Tok $\downarrow$} & \textbf{Eff $\uparrow$} & \textbf{HV $\uparrow$} & \textbf{IGD $\downarrow$} & \textbf{$|\mathcal{F}|$ $\uparrow$} \\
\midrule
Initial & 54.0 & 54.0 & 607 & 8.9 & n/a & n/a & 1 \\
{\WPRO}$_{0.3}$ & 56.0 & 55.8 & \textbf{239} & \textbf{23.4} & 54k & 0.46 & \textbf{2} \\
{\WPRO}$_{0.5}$ & 57.5 & 57.2 & 296 & 19.4 & 57k & 0.47 & \textbf{2} \\
{\WPRO}$_{0.7}$ & 61.5 & 60.1 & 419 & 14.7 & 58k & \textbf{0.29} & \textbf{2} \\
\SCULPT & 61.0 & 60.5 & 1{,}607 & 3.8 & 60k & 0.75 & \textbf{2} \\
\Distill & 55.0 & 55.0 & 350 & 15.7 & 55k & 0.45 & \textbf{2} \\
\LLMLingua & 55.0 & 55.0 & 497 & 11.1 & 52k & 0.46 & 1 \\
\CRAFT & \textbf{62.0} & \textbf{61.3} & 613 & 10.1 & \textbf{61k} & 0.55 & \textbf{2} \\
\midrule
\multicolumn{8}{c}{\textbf{GoE}} \\
\midrule
Initial & 34.0 & 34.0 & 794 & 4.3 & n/a & n/a & 1 \\
{\WPRO}$_{0.3}$ & 31.5 & 31.5 & $\infty$ & $\infty$ & 31k & 0.70 & \textbf{2} \\
{\WPRO}$_{0.5}$ & \textbf{43.0} & \textbf{40.0} & 856 & 5.0 & \textbf{40k} & \textbf{0.12} & \textbf{2} \\
{\WPRO}$_{0.7}$ & 39.5 & 39.2 & 845 & 4.7 & 38k & 0.39 & \textbf{2} \\
\SCULPT & 37.0 & 36.0 & 2{,}590 & 1.4 & 35k & 0.84 & \textbf{2} \\
\Distill & 33.0 & 33.0 & \textbf{389} & \textbf{8.5} & 32k & 0.42 & 1 \\
\LLMLingua & 34.0 & 33.0 & 581 & 5.9 & 33k & 0.34 & \textbf{2} \\
\CRAFT & 37.5 & 35.5 & 484 & 7.7 & 36k & 0.23 & \textbf{2} \\
\midrule
\multicolumn{8}{c}{\textbf{DQA}} \\
\midrule
Initial & 76.6 & 76.6 & 512 & 15.0 & n/a & n/a & 1 \\
{\WPRO}$_{0.3}$ & 73.7 & 70.5 & \textbf{190} & \textbf{38.8} & 74k & 0.41 & 2 \\
{\WPRO}$_{0.5}$ & 76.0 & 73.6 & 251 & 30.3 & 72k & 0.47 & \textbf{3} \\
{\WPRO}$_{0.7}$ & 70.3 & 68.5 & 228 & 30.8 & 70k & 0.65 & 2 \\
\SCULPT & \textbf{82.3} & \textbf{81.1} & 1{,}240 & 6.6 & \textbf{78k} & 0.52 & 2 \\
\Distill & 76.6 & 74.6 & 224 & 34.2 & 73k & 0.41 & 2 \\
\LLMLingua & 76.0 & 74.6 & 298 & 25.5 & 75k & 0.40 & \textbf{3} \\
\CRAFT & 77.7 & 75.1 & 225 & 34.5 & 77k & \textbf{0.20} & \textbf{3} \\
\bottomrule
\end{tabular}
\end{minipage}\hfill
\begin{minipage}[t]{0.49\textwidth}\centering
\begin{tabular}{lrrrrrrr}
\toprule
\multicolumn{8}{c}{\textbf{CJ}} \\
\midrule
\textbf{Method} & \textbf{Best $\uparrow$} & \textbf{Mean $\uparrow$} & \textbf{Tok $\downarrow$} & \textbf{Eff $\uparrow$} & \textbf{HV $\uparrow$} & \textbf{IGD $\downarrow$} & \textbf{$|\mathcal{F}|$ $\uparrow$} \\
\midrule
Initial & 68.5 & 68.5 & 499 & 13.7 & n/a & n/a & 1 \\
{\WPRO}$_{0.3}$ & 65.5 & 64.5 & \textbf{103} & \textbf{63.6} & 65k & 0.68 & 2 \\
{\WPRO}$_{0.5}$ & 63.1 & 62.0 & $\infty$ & $\infty$ & 62k & 0.78 & 2 \\
{\WPRO}$_{0.7}$ & 74.2 & \textbf{73.2} & 682 & 10.9 & 73k & \textbf{0.30} & 2 \\
\SCULPT & \textbf{76.9} & 72.7 & 1{,}784 & 4.3 & \textbf{75k} & 0.37 & \textbf{4} \\
\Distill & 68.5 & 68.5 & 140 & 48.9 & 63k & 0.47 & 1 \\
\LLMLingua & 65.2 & 65.1 & $\infty$ & $\infty$ & 65k & 0.62 & 1 \\
\CRAFT & 74.2 & 71.2 & 351 & 21.1 & 73k & \textbf{0.30} & 2 \\
\midrule
\multicolumn{8}{c}{\textbf{FF}} \\
\midrule
Initial & 71.8 & 71.8 & 526 & 13.7 & n/a & n/a & 1 \\
{\WPRO}$_{0.3}$ & 79.7 & 77.4 & \textbf{161} & \textbf{49.5} & 78k & 0.32 & \textbf{2} \\
{\WPRO}$_{0.5}$ & 80.7 & 80.3 & 204 & 39.6 & 74k & 0.44 & \textbf{2} \\
{\WPRO}$_{0.7}$ & 74.7 & 73.2 & 206 & 36.3 & 74k & 0.53 & \textbf{2} \\
\SCULPT & 73.9 & 72.2 & 1{,}219 & 6.1 & 69k & 1.06 & 1 \\
\Distill & 91.4 & 90.2 & 198 & 46.2 & 85k & \textbf{0.25} & \textbf{2} \\
\LLMLingua & \textbf{98.9} & \textbf{98.9} & 254 & 38.9 & \textbf{92k} & 0.46 & 1 \\
\CRAFT & 90.4 & 87.5 & 245 & 36.9 & 80k & 0.39 & \textbf{2} \\
\midrule
\multicolumn{8}{c}{\textbf{ST}} \\
\midrule
Initial & 62.1 & 62.1 & 419 & 14.8 & n/a & n/a & 1 \\
{\WPRO}$_{0.3}$ & 60.4 & 59.7 & \textbf{82} & \textbf{73.7} & 57k & 0.49 & \textbf{2} \\
{\WPRO}$_{0.5}$ & 61.8 & 59.5 & 98 & 63.1 & 58k & 0.44 & \textbf{2} \\
{\WPRO}$_{0.7}$ & 67.0 & 66.0 & 238 & 28.1 & 60k & \textbf{0.34} & \textbf{2} \\
\SCULPT & \textbf{71.7} & \textbf{71.7} & 1{,}290 & 5.6 & \textbf{70k} & 0.85 & \textbf{2} \\
\Distill & 61.2 & 60.9 & 149 & 41.1 & 61k & 0.46 & \textbf{2} \\
\LLMLingua & 60.0 & 59.0 & 419 & 14.3 & 60k & 0.56 & \textbf{2} \\
\CRAFT & 68.4 & 67.7 & 190 & 36.0 & 67k & \textbf{0.34} & \textbf{2} \\
\bottomrule
\end{tabular}
\end{minipage}
\caption{Main GPT-5 comparison at $R=8$ across the six datasets (left: BT, GoE, DQA; right: CJ, FF, ST). Bold marks the best per (dataset, column); lower is better for Tok and IGD, higher otherwise. ``n/a'' on the initial-prompt row marks HV and IGD that are undefined for a single point. ``$\infty$'' marks rows where no retained prompt clears the feasibility filter ($p \ge 0.95\,p_{\mathrm{init}}$); the marker applies to both Tok and Eff, which share the feasibility filter and are therefore both undefined in that case. Values averaged across two seeds.}
\label{tab:full_results}
\end{table*}

\begin{figure}[!t]
\centering
\includegraphics[width=\linewidth]{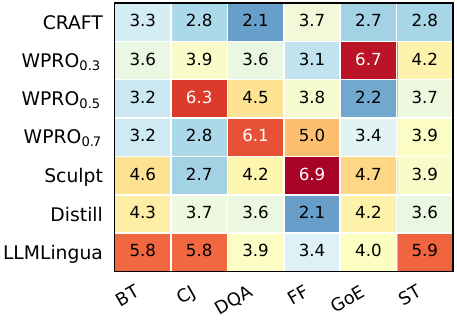}
\caption{Per-dataset composite rank at $R=8$ (rank $\downarrow$). \CRAFT leads DQA and ST; {\WPRO}$_{0.5}$ leads BT and GoE; \SCULPT leads CJ; \Distill leads FF.}
\label{fig:composite_rank_heatmap}
\end{figure}

\paragraph{Per-Dataset Detail.}
Across the six datasets, \CRAFT places 1st on DQA and ST, 2nd on CJ (tied with {\WPRO}$_{0.7}$) and GoE, 3rd on BT (within $0.11$ of the leader {\WPRO}$_{0.5}$), and 4th on FF. No other method matches this consistency: \SCULPT reaches rank $6.94$ on FF, {\WPRO}$_{0.3}$ reaches $6.72$ on GoE, {\WPRO}$_{0.7}$ reaches $6.06$ on DQA, {\WPRO}$_{0.5}$ reaches $6.28$ on CJ, and \LLMLingua reaches $5.89$ on ST. \CRAFT is the only method whose worst-case per-dataset rank stays below 4 (at $3.72$ on FF), reflecting robustness across the front-quality, score, and cost aspects rather than dominance on individual datasets.

\subsection{Per-Dataset Optimization Dynamics}
\label{app:dynamics}

\begin{figure*}[!t]
\centering
\includegraphics[width=\linewidth]{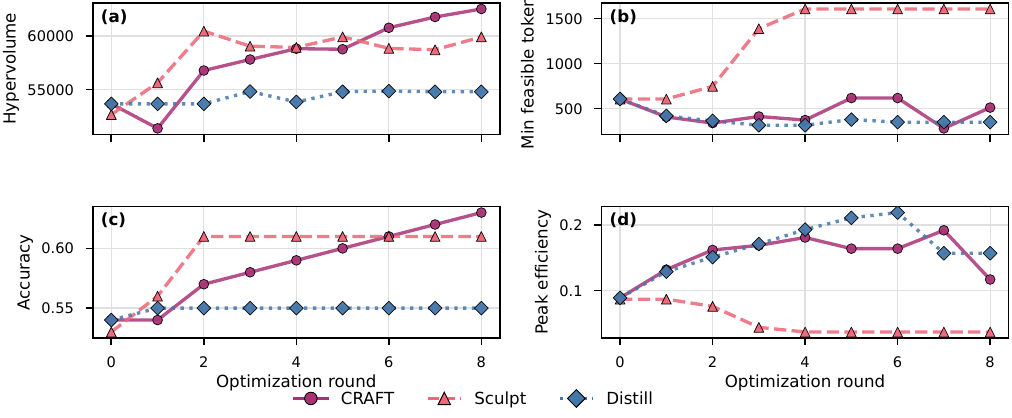}
\caption{Optimization dynamics on BT (GPT-5) over rounds 0 to 8 for \CRAFT, \SCULPT, and \Distill: (a) hypervolume, (b) min feasible tokens, (c) best score, (d) peak efficiency.}
\label{fig:metrics_grid}
\end{figure*}

\begin{figure*}[!t]
\centering
\includegraphics[width=\linewidth]{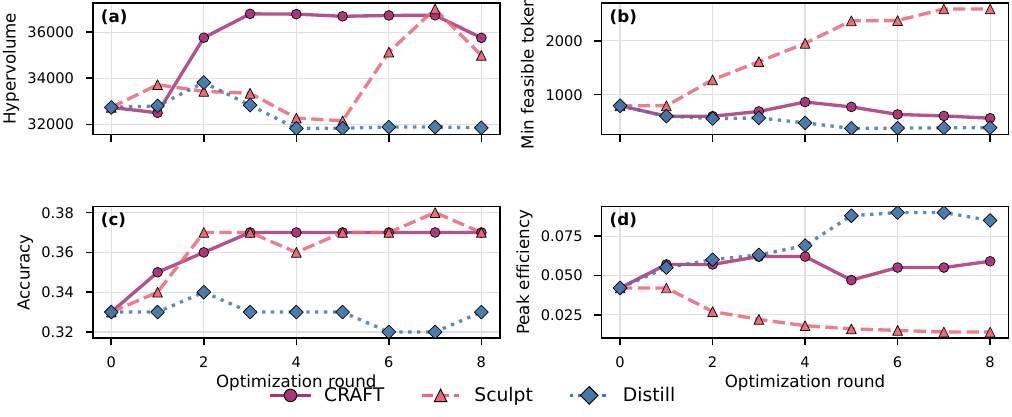}
\caption{Optimization dynamics on GoE (GPT-5) over rounds 0 to 8 for \CRAFT, \SCULPT, and \Distill: (a) hypervolume, (b) min feasible tokens, (c) best score, (d) peak efficiency.}
\label{fig:dyn_goe}
\end{figure*}

\begin{figure*}[!t]
\centering
\includegraphics[width=\linewidth]{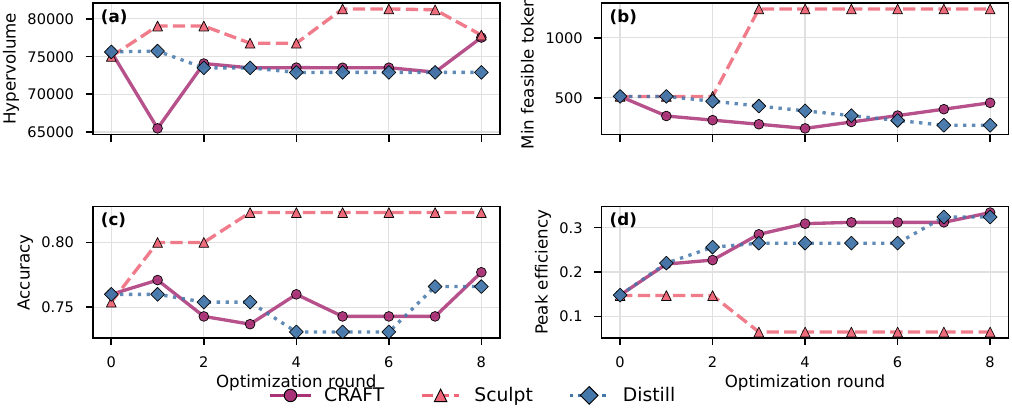}
\caption{Optimization dynamics on DQA (GPT-5) over rounds 0 to 8 for \CRAFT, \SCULPT, and \Distill: (a) hypervolume, (b) min feasible tokens, (c) best score, (d) peak efficiency.}
\label{fig:dyn_dqa}
\end{figure*}

\begin{figure*}[!t]
\centering
\includegraphics[width=\linewidth]{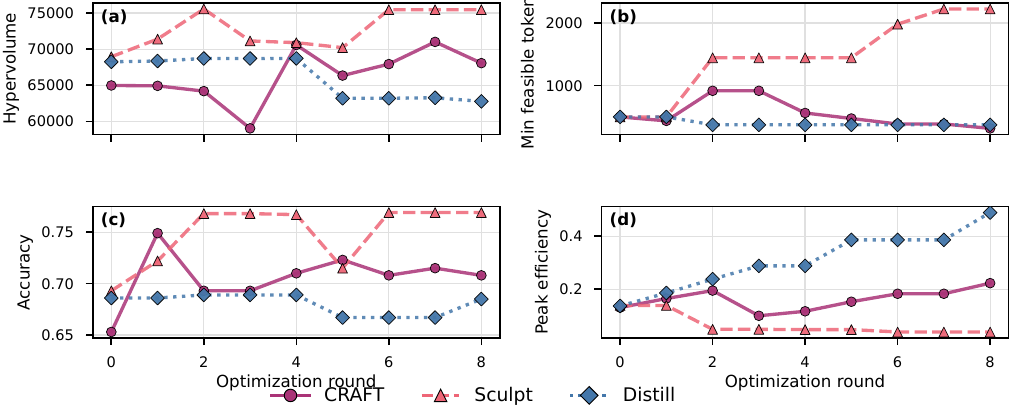}
\caption{Optimization dynamics on CJ (GPT-5) over rounds 0 to 8 for \CRAFT, \SCULPT, and \Distill: (a) hypervolume, (b) min feasible tokens, (c) best score, (d) peak efficiency.}
\label{fig:dyn_cj}
\end{figure*}

\begin{figure*}[!t]
\centering
\includegraphics[width=\linewidth]{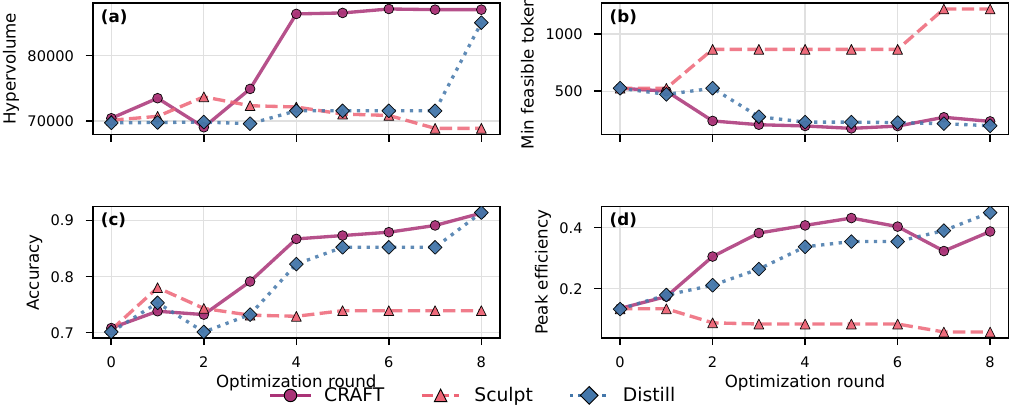}
\caption{Optimization dynamics on FF (GPT-5) over rounds 0 to 8 for \CRAFT, \SCULPT, and \Distill: (a) hypervolume, (b) min feasible tokens, (c) best score, (d) peak efficiency.}
\label{fig:dyn_ff}
\end{figure*}

\begin{figure*}[!t]
\centering
\includegraphics[width=\linewidth]{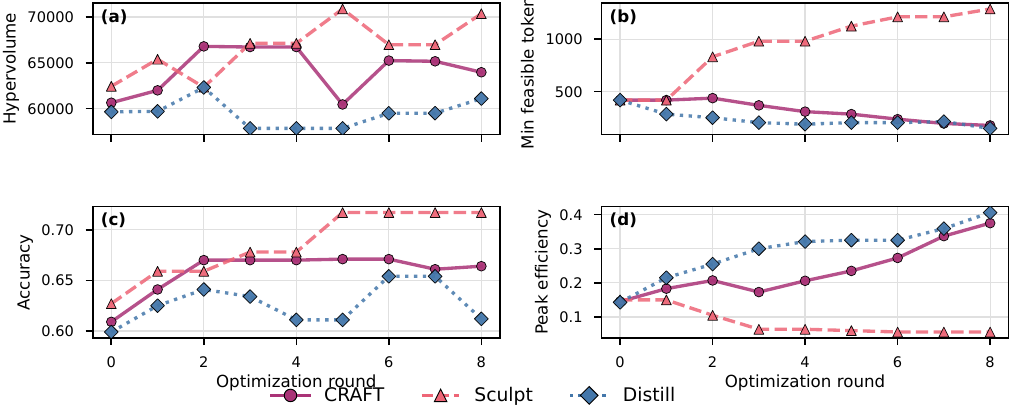}
\caption{Optimization dynamics on ST (GPT-5) over rounds 0 to 8 for \CRAFT, \SCULPT, and \Distill: (a) hypervolume, (b) min feasible tokens, (c) best score, (d) peak efficiency.}
\label{fig:dyn_st}
\end{figure*}

Figures~\ref{fig:metrics_grid} to \ref{fig:dyn_st} report the per-round dynamics view (hypervolume, min feasible tokens, best score, peak efficiency) for all six GPT~5 benchmarks, capped at the default $R=8$ snapshot. Gaps in the (b) Min feasible tokens curve mark rounds where no \Distill candidate passed the feasibility filter ($p \ge 0.95\,p_{\mathrm{init}}$); the other curves plot directly from each method's per-round Pareto report.

\section{Artifacts, Reproducibility, and Responsible Use}
\label{app:checklist}

\paragraph{Datasets, Models, and Code.}
We evaluate on six public benchmarks: BeaverTails \citep{ji2023beavertails}, GoEmotions \citep{demszky-etal-2020-goemotions}, and four BIG-Bench Hard tasks \citep{suzgun-etal-2023-challenging} (Disambiguation QA, Causal Judgement, Formal Fallacies, and Salient Translation Error Detection). The optimizer and target models are hosted LLMs: the GPT-5 family \citep{singh2025openai}, DeepSeek-V4 Flash \citep{deepseekai2026deepseekv4}, and Kimi-K2.5 \citep{team2026kimi}. The refiner and condenser build on \SCULPT \citep{kumar-etal-2025-sculpt}; baselines additionally use \OPRO \citep{yang2023large}, \Evo \citep{guo2024connecting}, and \LLMLingua-2 \citep{pan-etal-2024-llmlingua}. Every dataset, model, and baseline method is cited at first use. All six datasets are public research benchmarks for text classification or reasoning, and we use each only to evaluate prompt optimization, consistent with its intended research use.

\paragraph{Licenses.}
The datasets are distributed for research use: BeaverTails under CC~BY-NC~4.0, GoEmotions under Apache-2.0, and BIG-Bench Hard under the MIT license. Our use is non-commercial academic research and falls within these terms. The hosted LLMs are accessed through their providers' inference APIs under the respective API terms of service.

\paragraph{Computational Budget and Infrastructure.}
All experiments run through hosted LLM inference APIs; we train no model, so no GPU hours are consumed, and parameter counts for the API models are not publicly disclosed by their providers. The dominant cost is the number of LLM calls. Each round, \CRAFT issues $k(n_R+n_C)=48$ candidate-generation calls and $B\cdot|\mathcal{V}_{\pi(t)}|$ validation calls, with validation-subset sizes following Table~\ref{tab:dataset_stats}, repeated over $R=8$ rounds; the same per-round call budget applies to every compared method (\S\ref{sec:experiments}). Sentence-embedding and $k$-means clustering for the validation subsets run once per dataset on CPU and are negligible next to the LLM calls.

\paragraph{Software and Package Settings.}
The implementation is in Python. LLM calls are issued through the official \texttt{openai} and \texttt{azure-ai-inference} client libraries, and runs are configured with Hydra and OmegaConf. Prompt-token cost $c(P)$ is measured with the \texttt{tiktoken} tokenizer, and prompts are parsed and edited as trees with \texttt{networkx}. Validation subsets are constructed with $k$-means clustering \citep{macqueen1967kmeans} from \texttt{scikit-learn}, over \texttt{all-MiniLM-L6-v2} sentence embeddings \citep{reimers-gurevych-2019-sentence} produced by \texttt{sentence-transformers} on a PyTorch backend, using $T=8$ clusters. The \LLMLingua-2 baseline uses the \texttt{llmlingua} package \citep{pan-etal-2024-llmlingua} with its default token-level compression configuration. Numerical routines use \texttt{numpy}, \texttt{scipy}, and \texttt{pandas}. All LLM calls fix the generation temperature to $0$ (\S\ref{sec:experiments}). A full dependency list will be released with the code.

\paragraph{Potential Risks.}
\CRAFT optimizes whatever task score it is given. If that score rewards adversarial behavior (for example, eliciting unsafe outputs from the target LLM), the same loop that finds high-accuracy, low-cost prompts can also produce shorter and more reliable jailbreak prompts. Mitigation is the responsibility of whoever defines the task score; \CRAFT itself adds no safety constraints beyond those of the target LLM.

\end{document}